\documentclass{article}
\usepackage[utf8]{inputenc}
\usepackage{fullpage}
\usepackage{enumitem}
\usepackage{booktabs}
\usepackage{multirow}
\usepackage{natbib}
\usepackage{subcaption}
\usepackage{graphicx}
\usepackage{amsmath}
\usepackage{amsfonts}
\usepackage{amssymb}
\usepackage{nicefrac}
\usepackage{xcolor}
\usepackage{hyperref}
\usepackage{csquotes}
\usepackage{mdframed}
\usepackage{tablefootnote}
\usepackage{authblk}
\usepackage{float}
\usepackage{placeins}
\usepackage{tabularx}
\usepackage{pifont}
\usepackage{listings} 
\usepackage{tcolorbox}
\usepackage{diagbox}
\tcbuselibrary{breakable}

\newcommand{\gemini}{\texttt{Gemini 2.5 Pro}}
\newcommand{\gpt}{\texttt{GPT 5}}
\newcommand{\grok}{\texttt{Grok 4}}
\newcommand{\deepseek}{\texttt{DeepSeek Reasoner v3.1}}
\newcommand{\claude}{\texttt{Claude Sonnet 4.5}}

\newcommand{\benchmark}{FLAWS}

\title{\benchmark{}: A Benchmark for Error Identification\\ and Localization in Scientific Papers}

\author[1,2]{Sarina Xi\thanks{Work done during time at Carnegie Mellon University.}}
\author[2]{Vishisht Rao}
\author[2]{Justin Payan}
\author[2]{Nihar B. Shah}

\affil[1]{University of Wisconsin--Madison}
\affil[2]{Carnegie Mellon University}

\affil[1]{\texttt{sarina.xi@cs.wisc.edu}}
\affil[2]{\texttt{\{vsrao, jpayan, nihars\}@cs.cmu.edu}}

\date{}
\begin{document}
\maketitle
\begin{abstract}
The identification and localization of errors is a core task in peer review, yet the exponential growth of scientific output has made it increasingly difficult for human reviewers to reliably detect errors given the limited pool of experts. Recent advances in Large Language Models (LLMs) have sparked interest in their potential to support such evaluation tasks, from academic peer review to automated scientific assessment. However, despite the growing use of LLMs in review systems, their capabilities to pinpoint errors remain underexplored. In this work, we introduce Fault Localization Across Writing in Science (\benchmark{}), an automated benchmark consisting of 713 paper-error pairs designed to evaluate how effectively LLMs detect errors that undermine key claims in research papers. We construct the benchmark by systematically inserting claim-invalidating errors into peer-reviewed papers using LLMs, paired with an automated evaluation metric that measures whether models can identify and localize these errors. Developing such a benchmark presents unique challenges that we overcome: ensuring that the inserted errors are well-defined, challenging, and relevant to the content of the paper, avoiding artifacts that would make identification trivial, and designing a scalable, automated evaluation metric. On the resulting benchmark, we evaluate five frontier LLMs: \claude{}, \deepseek{}, \gemini{}, \gpt{}, and \grok{}. Among these, \gpt{} is the top-performing model, achieving 39.1\% identification accuracy when $k=10$, where $k$ is the number of top-ranked error text candidates generated by the LLM.
\end{abstract}

\section{Introduction}
Peer review underpins modern research across various disciplines. By subjecting research to evaluation by qualified experts in the field, peer review serves as a key quality control mechanism for ensuring rigor, soundness, and originality in published work. However, despite its central role in academia, peer review faces well-documented challenges --- including inconsistency, miscalibration, and bias --- that undermine its reliability \citep{shah2021}. Controlled studies have shown that human expert reviewers may fail to detect major errors in papers and still recommend acceptance for flawed papers \citep[Section 10.1]{shah2021}. Such oversights can have lasting consequences: erroneous publications may continue to shape subsequent research even after corrections or retractions \citep{shepperd2023}, while retractions themselves impose substantial financial and scientific costs \citep{stern2014}. These issues are exacerbated by the exponential growth of scientific output \citep{landhuis2016}. In Machine Learning (ML) and Artificial Intelligence (AI), submissions to leading conferences have increased more than tenfold over the past decade \citep{nip2024,iclr2025}, while the pool of qualified reviewers remains limited. As a result, identifying methodological flaws and factual errors before publication has become increasingly critical yet more difficult. 

To address these challenges, there has been growing interest in leveraging Natural Language Processing (NLP) to support aspects of the peer-review process \citep{kuznetsov2024}. In particular, Large Language Models (LLMs) are increasingly being explored to assist reviewers or even generate complete reviews. Reflecting this trend, AAAI 2026, a major AI conference, has incorporated LLM-generated reviews to provide “an additional perspective alongside traditional human expert evaluations” \citep{aaai_genai}. Research in this space ranges from early exploratory studies \citep{liu2023} to LLM-based automated reviewing frameworks \citep{yu2024, darcy2024, tyser2024, idahl2025}. Considering the widespread usage of LLM reviewers, several works have aimed to evaluate the ability of AI, and LLMs in particular, to produce high quality reviews. Some works evaluate quality based on how accurate AI reviewers are in predicting past human review scores \citep{checco2021aiassisted, tyser2024, idahl2025}, while others use human evaluations for assessing review quality \citep{liang2023usefulfeedback, darcy2024}. However, human reviews and evaluations themselves can contain bias, miscalibration, and subjectivity serving as a baseline. 

In this work, rather than full review generation, we focus on error identification and localization, a critical subtask of peer review that is often time-consuming and cognitively demanding for human reviewers. Our approach is motivated by scenarios of human reviewers using LLMs to detect errors in papers. LLMs will surface and highlight potential errors in manuscripts, guiding  human reviewers to where possible problems lie. The human reviewers can then focus their attention and expertise to manually verify these potentially problematic parts of the paper. Recent work highlights the potential of frontier LLMs to detect and correct objective mistakes \citep{bianchi2025}, demonstrating the promise of using LLMs for error identification. To accomplish this vision, we build our benchmark and evaluation metrics to assess LLMs' abilities to identify the text containing errors, and also ask LLMs to rank the output texts according to the chances of containing a major error.

Existing methods of eliciting reviews from LLMs have not been designed or evaluated for the error identification and localization subtask. Prior work has shown that LLMs, such as GPT-4, struggled to identify flaws in $13$ papers when asked to ``review the paper" holistically, but performed better when specifically prompted to detect errors \citep{liu2023}. This pattern mirrors findings from human reviewers, where evaluating multiple criteria simultaneously can reduce the quality of assessment for each individual criterion \citep{lane2024}. Furthermore, holistic reviews or asking LLMs to generate review scores may not reveal the actual locations of errors, providing less actionable guidance for human reviewers to identify errors in papers. By isolating error identification and localization from broader review tasks, we can more precisely evaluate and potentially improve LLM capabilities in this high-stakes subtask of peer review.

Given this motivation, it becomes crucial to systematically evaluate LLMs' capabilities for error identification and localization. To contextualize our work better, we now summarize the works that are most relevant to this setting; we discuss them in more detail in Section \ref{section:related_work}. \citet{zhang2025} study the identification of critical errors using LLMs. However, their work is based on WithdrarXiv \citep{rao2024}, a set of retracted arXiv papers and retraction notes. They acknowledge that the retraction notes often provide ambiguous reasons and may not specify the exact error, undermining evaluation reliability. \citet{son2025} build on the same dataset and obtain 91 human-annotated errors for 83 papers. This approach provides definitive error labels, but is costly and difficult to scale. Similarly, \citet{liu2023} manually construct 13 papers each with a deliberately inserted error and evaluate LLMs on their ability to correctly find the errors. To avoid extensive manual annotation while having well-defined errors, \citet{xu2025llm} curate a synthetic benchmark. However, their focus is on identifying and addressing \emph{limitations}, which include non-fatal flaws such as irrelevant citations or lack of thorough ablations, rather than isolating the task of fatal error identification. In addition, they used a predefined set of limitations and noted that while this ensures fine-grained control, this means that their approach does not generalize outside of the 11 limitations they studied. There is currently no scalable and generalizable benchmark to systematically evaluate LLMs on error identification in research papers.

To address this gap, we introduce Fault Localization Across Writing in Science (\benchmark{}), an automated benchmark that inserts controlled modifications to invalidate the central claims of papers. Our approach systematically introduces paper-specific errors in an explicit, generalizable, and scalable manner. By focusing specifically on error identification and localization rather than full review generation, our benchmark isolates a clearly defined and critical component of the peer-review process, helping human reviewers concentrate their limited time and attention on the most problematic text excerpts in paper submissions. Overall, our contributions are threefold:

\begin{itemize}
    \item \textbf{Framework for controlled error insertion}: We introduce a systematic framework for inserting errors invalidating the key claims of peer-reviewed papers. This approach addresses the challenge of introducing errors that are directly relevant to the scientific content of a paper while maintaining clear and explicit error labels. Furthermore, it can be applied to any paper without being limited to a rigid set of predefined error categories. The pipeline also filters out trivial or superficial edits, yielding a generalizable and well-defined testbed for evaluating the ability of an LLM to detect and locate substantive, challenging errors.
    \item \textbf{Automated benchmark and evaluation}: We present an automated benchmark, \benchmark{}, comprising 713 paper-error pairs. These pairs come from 448 unique papers with errors inserted by \gemini{} and 265 unique papers with errors inserted by \gpt{}. Our carefully designed evaluation metric enables a scalable assessment of the ability of LLMs to identify and localize errors without requiring expensive manual annotation.
    \item \textbf{Evaluation of state-of-the-art LLMs}: We evaluate five leading LLMs on our benchmark and provide a ranking of their ability to identify and localize errors in scientific papers, as shown in Table \ref{tab:model_ranking}. Among these five models, \gpt{} stands out as the top-performing model, achieving 39.1\% identification accuracy when $k=10$, where $k$ is the number of top-ranked error text candidates generated by the model.

    \begin{table}[th!]
    \centering
    \begin{tabular}{lcccc}
    \hline 
    \multirow{2}{*}{\textbf{Identification Model}} & \multirow{2}{*}{\textbf{Rank}} & \multirow{2}{*}{\textbf{Score $\beta_j$}} & \multicolumn{2}{c}{\textbf{Accuracy}}\\
     &  & & \textbf{@k=3} & \textbf{@k=10}\\
    \hline
    \gpt{} & 1 & 2.10 & 19.2\% & 39.1\%\\
    \deepseek{} & 2 & 1.90 & 16.3\%& 35.2\%\\
    \grok{} & 3 & 1.68 & 16.3\% & 23.4\%\\
    \claude{} & 4 & 1.47 & 12.6\% & 21.5\%\\
    \gemini{} & 5 & 1.41 & 15.7\% & 19.8\%\\
    \hline
    \end{tabular}
    \caption{Ranking of LLM performance on \benchmark. The ``Accuracy'' columns correspond to the fraction of instances where the top-ranked $k$ error candidate texts output by the LLM actually contain the error. To mitigate any confounding factors and evaluate overall performance, we perform analysis on the identification results and use the calculated score, $\beta_j$, to rank the models (see details in Section \ref{section:results}).}
    \label{tab:model_ranking}
\end{table}
\end{itemize}
We release our code and benchmark dataset at \url{https://github.com/xasayi/FLAWS}.

\section{Related Work}
\label{section:related_work}
In this section, we first review current applications of LLMs in peer review. Then, we examine existing benchmarks that are relevant to error identification in academic papers using LLMs.

\paragraph{LLMs for peer review automation.} Research on applying LLMs to the peer-review process has expanded rapidly. Early work looked at whether LLMs can be reviewers or help with parts of the review pipeline. Studies such as \citet{yuan2022} and \citet{liu2023} found that while LLMs can provide surface-level insights, they fall short of producing reviews that meet expert standards. A key finding from \citet{liu2023} also showed a core insight: LLMs were poor at finding flaws when asked to review the paper holistically, but did better when prompted specifically to find errors. Despite these early limitations, later research shifted to automating the generation of full review reports. For example, \citet{yu2024} proposed an LLM-based framework to automatically generate reviews, while \citet{darcy2024} used multiple LLMs that engage in internal discussions to produce reviews.

The rapid growth of LLM review generation systems made it necessary to evaluate their quality and reliability. One line of work studied how well AI reviewers predicted past human review scores \citep{checco2021aiassisted,idahl2025,shcherbiak2024evaluating,thelwall2024predictive}. The drawback to this approach is that past review scores themselves contain bias, subjectivity, and miscalibration. Another line of work uses subjective human evaluations to annotate these reviews \citep{liang2023usefulfeedback, darcy2024, tyser2024}. Recent work also explores using gold-standard reference-free evaluation of LLM generated reviews \citep{xu2025}. However, they still rely on human-generated content as input. \citet{goldberg2025peer} conducted a randomized controlled trial and asked other reviewers, meta-reviewers, and experts to evaluate reviews. Then, they asked the authors to evaluate reviews for their own paper and showed that human evaluations can contain various biases such as increased focus on style over substance. 

Several works found major failures in the quality of LLM-generated reviews. For example, \citet{shin2025} found that LLM reviews have biased blind spots, focusing on technical validity while significantly overlooking novelty (though the authors did not evaluate the correctness of the technical assessments). \citet{zhou2024} identified persistent weaknesses in processing long papers and in providing critical feedback. Similarly, \citet{ye2024we} found that LLMs have serious inherent flaws, such as being easily manipulated by authors, favoring papers from well-known authors, and lacking independent critical judgment. Systematic comparisons by \citet{du2024} showed that LLMs produce more ``deficient review segments" as well as superficial critiques that are ``paper-unspecific" and often fall outside the paper's scope. \citet{garg2025} introduced an evaluation framework called ReviewEval which highlighted the lack of actionable insights, analytical depth, and adherence to review guidelines in LLM-generated reviews. To overcome some of these flaws, \citet{idahl2025} fine-tuned a special model, OpenReviewer, to combat the overly positive nature of reviews produced by general-purpose LLMs and to better match the critical ratings from human reviewers. These findings all support the view of \citet{seghier2024ai}, who argues that AI-powered review has significant ethical concerns and must have human supervision rather than replacing human judgment.

These demonstrated failures of the holistic review approach caused a pivot toward more focused, fine-grained tasks. One direction is automated scientific feedback. Systems developed by \citet{weng2025} train LLM-based reviewer agents to support automated research workflows, and \citet{chamoun2024} propose feedback tools to help authors improve their scientific writing. More critically for our work, the unreliability of holistic reviews has also pushed research to check scientific validity directly. The prior evaluation studies show that LLMs diverge systematically from human reviewing tendencies, underscoring the need for more fine-grained evaluation of scientific content. Motivated by this, instead of emphasizing holistic review production, our focus is on the specific ability to identify errors.

\paragraph{Relevant datasets.} To study LLM behavior in peer-review-like settings, several datasets have been constructed. NLPEER \citep{dycke2023} provides one of the first large-scale corpora of papers paired with human reviews, enabling tasks such as predicting the review score.

Several works are more closely related to our goal of evaluating error identification, which fall into two main groups. The first group uses real-world papers that are already known to be flawed. For instance, \citet{zhang2025} use papers withdrawn from arXiv along with the author-submitted retraction comments via a dataset called WithdrarXiv \citep{rao2024}. \citet{son2025} also use this dataset, along with papers and their associated comments and discussions on PubPeer, an anonymous post-publication peer review website where users can flag errors/issues in papers and open up discussion. The main issue with using WithdrarXiv is that arXiv papers have not always been peer reviewed, hence, there may be errors besides the stated reason for retraction. In addition, \citet{zhang2025} acknowledge a limitation of using this dataset, in that retraction notes often do not actually specify the exact error. Papers appearing on PubPeer, on the other hand, can be subject to issues like misinterpretations. Finally, constructing benchmarks based on real-world retractions and comments often involves manual identification of errors, making it not as scalable as our approach, which introduces errors into papers automatically. 

To solve the problem of noisy data and scalability, a second group of datasets uses synthetic errors. \citet{xu2025llm} introduce synthetic ``limitations'' into experimental NLP papers. Their prompts are manually designed for specific limitation types.  Although ``limitations'' and ``errors'' exhibit some overlap, such as incorrect application of methods, limitations include other downsides that are not immediately fatal to a paper's arguments (i.e., would void the validity of the paper), e.g., oversights in the literature review or lack of thorough ablation studies. Moreover, they only applied their method to experimental NLP papers. In contrast, our error-insertion prompts are designed to insert fatal errors of any error type by modifying passages to break a claim made in the paper. Our approach is applicable to any paper, including purely theoretical and survey papers. \citet{dycke2025} insert synthetic counterfactual logical errors into papers, but their evaluation still asks the LLM for a review instead of explicitly asking it to find errors. They evaluate error-detection by studying how much the LLM reviews change their focus on the ``soundness'' aspect of the review, measured by a RoBERTa model. This approach does not explain how to evaluate LLMs for their ability to \emph{locate} the error-containing text.

\citet{tyser2024} and \citet{li2025} introduce limitations and errors into papers based on specific error types automatically using LLMs and ask an LLM to generate reviews for the paper before and after the modification. They study how scores of LLM-generated reviews change depending on the types of errors inserted. \citet{lou2025} target a narrow set of mistakes, focusing mostly on erroneous equations. Our work differentiates itself from these works, as our error insertion prompts are not tied to specific error type. More importantly, our method evaluates the ability of models to explicitly identify errors in the paper by localizing the errors to a certain passage in the paper. Rather than measuring the differences in review scores alone, our method provides a much more rigorous way to evaluate whether models are truly able to identify errors in articles when they are actively prompted to do so.

Finally, broader scientific reasoning benchmarks provide additional context. Some related benchmarks \citep{kamoi2024evaluating} focus on detecting errors in the LLM's own reasoning, rather than in external scientific papers. Other benchmarks are difficult and assess the reasoning abilities of LLMs, but have short contexts \citep{phan2025, ye2025aimepreview}. There are also long-context benchmarks \citep{ling2025, bai2025}, which assess extended document understanding but do not require models to judge the correctness of scientific arguments. Our work directly addresses these gaps by designing a benchmark to evaluate an LLM's ability to check scientific validity in diverse, long-context research papers.
 
\section{Benchmark Construction}
\begin{figure}[h]
    \centering
    \includegraphics[width=1.0\linewidth]{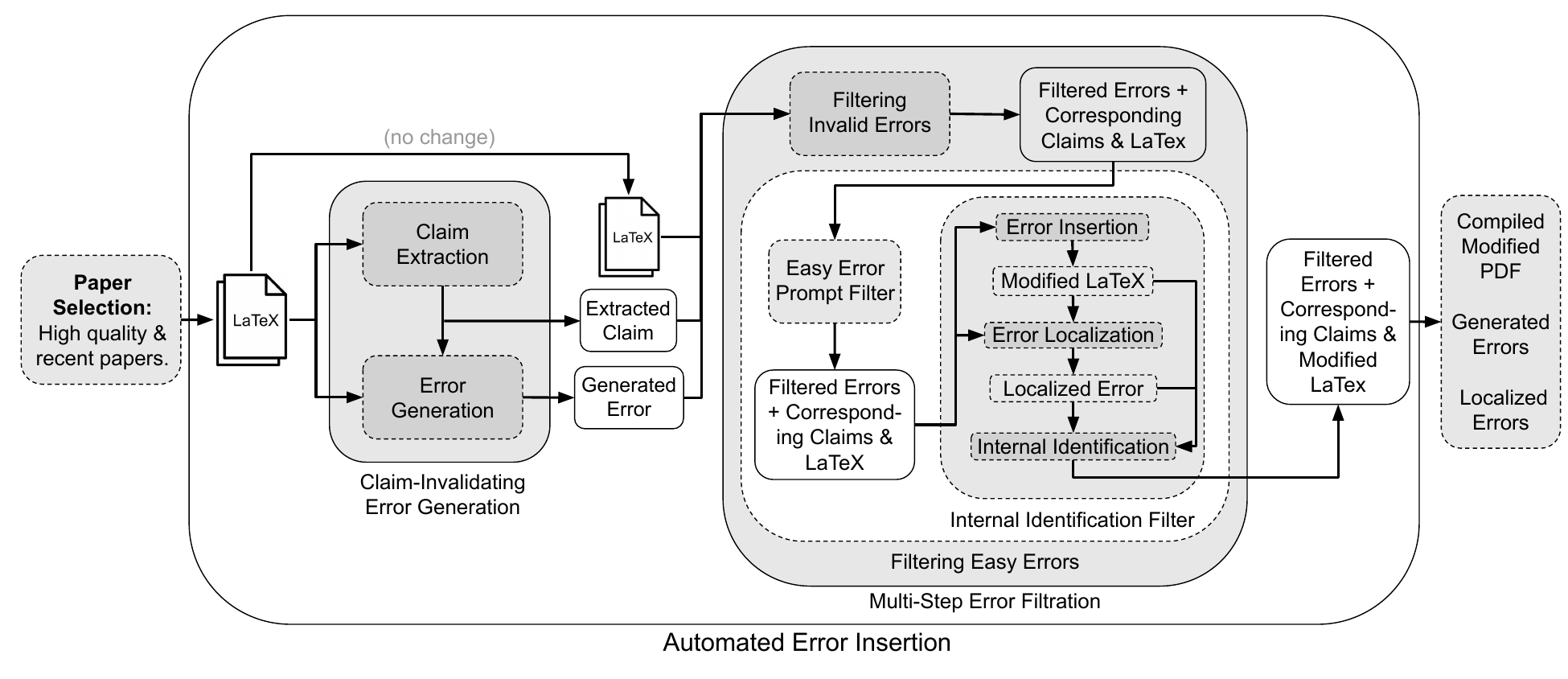}
    \caption{Detailed benchmark construction workflow.}
    \label{fig:benchmark_construction}
\end{figure}

In this section, we detail the construction of \benchmark{}, where the overall workflow is presented in Figure \ref{fig:benchmark_construction}. We begin by outlining how we carefully selected our paper corpus used as the foundation for our dataset. Then, we present our automated error insertion pipeline, which introduces high-quality errors in the papers by claim-invalidating error generation and multi-step error filtration. 
\subsection{Paper Selection}
To construct our dataset, we first select a set of papers into which synthetic errors will be inserted. These papers must satisfy three criteria:

\begin{itemize}
    \item \textbf{Quality}: The subset of papers selected should be of reasonably high quality. While peer-reviewed papers can still contain errors, the overall quality is considered to be generally higher than non-peer-reviewed work. If a paper already has a significant mistake, an LLM might correctly identify that error rather than the synthetic flaw, which could unfairly penalize the model for detecting a genuine issue instead of missing the inserted one.
    \item \textbf{Recency}: Papers must be sufficiently recent to reduce the chances that they appear in the pretraining data of the LLMs. If a model has previously seen a paper during pretraining, its outputs may be a result of memorization rather than genuine identification. By selecting recently posted papers, we minimize this risk of data contamination. The models we evaluate have the following stated knowledge cutoff dates:\footnote{We acknowledge that data leakage could still occur: Knowledge cutoffs typically apply to pretraining, and vendors do not guarantee that models were not post-trained on more recent data.}
    \begin{table}[th!]
        \centering
        \begin{tabular}{ll}
        \hline
            \textbf{Model} & \textbf{Knowledge Cutoff Date} \\
            \hline
            \deepseek{} & July 2024\\
            \gpt{} & September 30, 2024 \\
            \grok{} & November 2024 \\
            \gemini{} & January 2025 \\
            \claude{} \footnotemark & March 2025 \\
            \hline
        \end{tabular}
        \caption{Knowledge cutoff dates for models. }
        \label{tab:cutoff_dates}
    \end{table}
    \footnotetext{We initially experimented with \texttt{Claude Sonnet 3.7} (knowledge cutoff: Nov 2024), but it frequently failed to detect any errors. We therefore opted for \claude{}, which exhibited stronger identification performance. While there is possible data leakage due to the March 2025 knowledge cutoff date, \claude{} has poor performance as found in Section \ref{section:identification_results}.}
    \item \textbf{Format}: The LaTeX source of papers must be available. LLMs handle error insertion far more effectively when provided with the raw LaTeX source, whereas supplying PDFs requires the model provider to perform object character recognition, which can introduce additional noise and formatting artifacts. Working directly with LaTeX source files ensures cleaner inputs and enables more precise error insertion.
\end{itemize}
To satisfy these criteria, we selected papers accepted to the 2025 International Conference on Machine Learning (ICML), and which were first posted on arXiv after January 2025. Acceptance at this venue guarantees that each paper has undergone peer review, establishing a baseline level of quality. The recency of ICML 2025 also reduces the risk of data contamination, as these papers are less likely to have appeared in the training corpora of current LLMs. Using the OpenReview API, we retrieved metadata of accepted papers and matched their titles to arXiv entries via the arXiv API to identify those with available LaTeX sources. For matched papers, we downloaded the corresponding LaTeX packages from arXiv on October 9, 2025, obtaining \textbf{623} papers.

While most papers contain a single main \texttt{.tex} file encompassing the entire paper, some papers organize content across multiple \texttt{.tex} files (e.g., separate files for sections or appendices). In these cases, we merge all relevant files into a single compilable LaTeX source file, which is then provided to the LLM for controlled error insertion.

\subsection{Automated Error Insertion}
We design an automated pipeline to introduce controlled errors into scientific papers. The process consists of two main stages: claim-invalidating error generation and multi-step error filtration, which together ensure that the inserted errors are meaningful, non-trivial, and properly localized. Table \ref{tab:filtering_papers} shows the number of errors and papers at each step in the insertion pipeline. Errors that survive the filtering steps are inserted into the original LaTeX source, and the modified PDFs are compiled. To further enhance robustness and reduce model-specific bias, we use two separate insertion models, \gemini{} and \gpt{}, enabling comparisons between models and ensuring that error generation and identification are not reliant on the same LLM. In this section, we describe the detailed protocol for error insertion, with Section \ref{section:manual_annotation} showing the reliability of our inserted errors. A full example of the error insertion process is provided in Appendix \ref{appendix:example_error}.

\begin{table}[h]
    \centering
    \begin{tabular}{p{4.6cm}cccc}
    \hline
    
    \multirow{2}{*}{\textbf{Insertion Model}} & \multicolumn{2}{c}{\textbf{\gemini{}}}& \multicolumn{2}{c}{\textbf{\gpt{}}} \\

    & \# Papers & \# Errors & \# Papers & \# Errors\\
    \hline
     
    Claim Extraction & 623 & - & 623& - \\
    Error Generation & 623 (100\%) & 4025 & 623 (100\%)& 9154 \\
    \hline
    Filtering Invalid Errors & 623 (100\%) & 4008 (99\%) & 623 (100\%)&  9105 (99\%)\\
    Filtering Easy Errors Stage 1, \textit{Easy Error Prompt Filter} & 623 (100\%)& 3373 (84\%) & 552 (87\%) & 1722 (19\%)\\
    (Filtering Easy Errors Stage 2,
    \textit{Hallucination Filter}) & 573 (92\%) & 2971 (74\%) & 476 (76\%) & 1358 (15\%) \\
    Filtering Easy Errors Stage 2, \textit{Internal Identification Filter} & 526 (84\%)& 1560 (39\%) & 341 (58\%) & 632 (7\%)\\
    \hline
    PDF Compilation & 448 (72\%)& 1287 (32\%) & 265 (43\%)& 468 (5\%)\\
    \hline
    \end{tabular}
    \caption{Number of unique papers and errors after each step in the error insertion pipeline. The percentages are based on the initial number of papers and errors. }
    \label{tab:filtering_papers}
    
\end{table}

\subsubsection{Claim-Invalidating Error Generation}
\paragraph{Claim extraction.} 
Prior approaches to synthetic error insertion in scientific papers typically rely on a predefined set of error types. While convenient, fixed error types may not generalize across domains, and can cover only a subset of possible mistakes. In addition, predefined errors may also not align with the content of individual papers, leading to errors that may be disconnected from the core scientific arguments. In contrast, our approach focuses on introducing errors undermining the key claims within each paper such that the errors are meaningful and directly relevant to the paper's core content. This method reduces reliance on rigid error types and avoids the coverage gap that arises when the set of error types is incomplete.

To implement this, we identify claims that capture a paper's essential scientific contributions. We prompt an LLM with the LaTeX source of the paper and instruct it to generate a list of unique, falsifiable claims that reflect the main arguments or findings. These claims then serve as anchors for targeted error insertion. By grounding errors in the paper's core claims rather than arbitrary error types, we introduce mistakes that challenge models to reason about the substance of the research. The detailed prompt for claim generation is provided in Appendix \ref{appendix:claim_extraction}. 

\paragraph{Error generation.} For each generated claim in each paper, we introduce a single error that undermines it. We provide the LLM with both the claim and the corresponding LaTeX source, and instruct it to introduce a sophisticated, domain-appropriate error by modifying one or more segments of the text. During preliminary experiments, we observed that LLMs sometimes defaulted to producing superficial or stylistic mistakes. To address this, we refined the prompt to require that the inserted error represent a core conceptual flaw, remain logically coherent within the surrounding context, and be detectable from the text alone. The LLM is asked to output the modified text, the original text, and an explanation of the inserted error. To ensure the error is consistent and coherent, the model may modify multiple text excerpts in different parts of the paper to introduce a single error. Full prompting constraints and requirements are provided in Appendix \ref{appendix:error_generate}.

In practice, some models, particularly those with stricter safety guardrails, occasionally refuse to generate erroneous scientific content, citing policy restrictions. Although we apply targeted prompt engineering to mitigate this, we still encounter occasional refusals. These cases are discarded. If an LLM cannot generate any valid errors from any claims of a particular paper, we skip that paper entirely; however, this situation rarely occurs. With this particular set of 623 papers, we are able to successfully generate errors for each paper, with a total of 9154 errors generated by \gpt{} and 4025 errors generated by \gemini{} as indicated in Table \ref{tab:filtering_papers}. 

\subsubsection{Multi-Step Error Filtration}
\label{section:multi_step_filtration}

\paragraph{Filtering invalid errors.} To ensure that the generated errors are valid, we apply an LLM-based filter. We prompt a fresh instance of the error generation LLM to verify that the modified text excerpt (i) does not explicitly state that there is an error and (ii) does not simply surface an existing limitation in the paper. Specifically, we provide the LLM with the claim, the modified text excerpt, the error explanation, and the original LaTeX source, and ask whether the error satisfies the outlined criteria. Errors that fail this check are discarded. The detailed prompt is provided in Appendix \ref{appendix:invalid_error_filter}. As shown in Table \ref{tab:filtering_papers}, very few errors were considered as invalid.

\paragraph{Filtering easy errors.} We further refine our subset of errors by filtering out errors that are too easy using a two-step process.
\begin{enumerate}
    \item \textbf{Easy error prompt filter.} First, we prompt a fresh instance of the error generation LLM to determine if an error is too easy by providing it with the claim, the modified text excerpt, the error explanation, and the original LaTeX. The detailed prompt is provided in Appendix \ref{appendix:easy_error_filter}.
    \item \textbf{Internal Identification filter} Second, we use the same LLM for error generation to perform identification and discard errors that have been identified to further filter out easy errors; we call this process internal identification. Specifically, we insert an error into the LaTeX source to create a modified version, then prompt the LLM to identify errors from this modified source. To verify whether an error has been identified, we also perform error localization, which identifies all ground truth text excerpts corresponding to the inserted error, as the inserted error may affect excerpts that are not directly modified. Any errors that are internally identified are discarded. Detailed procedures for error insertion, error localization, and internal identification are provided below.
    \begin{enumerate}[label=\alph*.]
        \item \underline{Error insertion.} For each error that passes the previous filtering step, we insert the modified text excerpt into the original LaTeX source. One approach would be to ask the LLM to output the start and end indices of the spans it modified. However, recent work shows that LLMs cannot consistently output numerical text spans \citep{hasanain2024, kasner2025}. Instead, we locate the replacement span by first searching for an exact match of the original excerpt in the LaTeX source. Since LLMs may slightly alter tokens or hallucinate text, exact matches are not always found. In such cases, we use the Python package \textit{difflib.SequenceMatcher} \citep{difflib} to find the closest match and replace the corresponding span with the modified text.

        We considered alternatives such as edit distance, n-gram overlap, and embedding-based similarity. Edit distance captures overall similarity but does not provide localized alignment indices. N-gram or token overlap is brittle with formatting and punctuation variations. Embedding-based approaches tolerate paraphrasing but risk incorrect matches if multiple semantically similar passages exist. In contrast, \textit{SequenceMatcher} is robust to minor textual variations and provides a similarity score in $[0,1]$, where a value closer to 1 indicates higher similarity. We replace the original with the modified text if the similarity exceeds 0.9. Low similarity often indicates LLM hallucination, so the 0.9 threshold conservatively minimizes erroneous replacements. The errors remaining after this step corresponds to the \textit{Hallucination Filter} in Table \ref{tab:filtering_papers} .
        \item \underline{Error localization.}
        For paper-error pairs where the error is successfully inserted, while the modified excerpts generated by the LLM introduce an error, the actual error may extend beyond the modified passages. For example, altering a description of an algorithm's objective could implicitly affect its stated complexity. To identify all error locations, we provide a fresh instance of the error generation LLM with the LaTeX source, the original and modified excerpts, and the associated claim. Then, we prompt the model to return all excerpts of the paper that pertain to the inserted error. We merge these localized error excerpts with the original modified excerpts to better capture all relevant error passages, and use this combined set as the ground truth error excerpt set. The detailed prompt is in Appendix \ref{appendix:error_location}.
        \item \underline{Internal identification.} Finally, we perform internal identification using a fresh instance of the error generation LLM. The model receives the altered LaTeX source and is informed that an error has been introduced by breaking a claim. It is then asked to locate the corresponding error text excerpts. If the LLM successfully identifies the error, the example is discarded. The specific prompt used is outlined in Appendix \ref{appendix:self_identification}.

        To determine whether an error has been correctly identified, we compute a word-level Levenshtein similarity between the returned excerpts and ground truth error excerpt set; a higher score indicates more similarity between two text excerpts. The same metric is used in the evaluation process to see whether an LLM successfully identified an error (more details in Section \ref{section:evaluation}). An error is considered successfully identified, and therefore filtered, if the similarity exceeds 0.5. This threshold was calibrated manually on a small set of examples to align with human judgment of error presence. After this internal identification step, only 39\% of the \gemini{}-generated errors remain while only 7\% of the \gpt{}-generated errors remain. This leaves 526 papers in the \gemini{} subset and 341 papers in the \gpt{} subset. We note that while this process removes a substantial number of easy errors, the stochastic nature of LLM generations means that a small amount of additional errors could be filtered out if the process was repeated multiple times. However, as the filtration process is very costly and running it multiple times has diminishing gains, we run the filtration pipeline once.

    \end{enumerate}
\end{enumerate}
    
\subsubsection{PDF compilation.} To better approximate a realistic peer-review setting, we evaluate LLM error identification on compiled PDFs rather than LaTeX source files, as human reviewers typically read the PDF and review platforms rarely collect LaTeX source code. We compile each modified LaTeX source code into a PDF using pdflatex via Python. The LaTeX source of some paper-error pairs failed to compile due to issues such as missing assets or unresolved dependencies, thus we discard these paper-error pairs. After this step, we obtain 265 papers in the \gpt{} subset and 448 papers in the \gemini{} subset. For each paper in each subset, if there are multiple compiled error-modified PDFs, we randomly sample one of them. This yields a benchmark of 713 distinct paper-error pairs used to evaluate the identification LLMs. 

\section{Automated Evaluation Framework}
\label{section:evaluation}
\begin{figure}[h]
    \centering
    \includegraphics[width=0.8\linewidth]{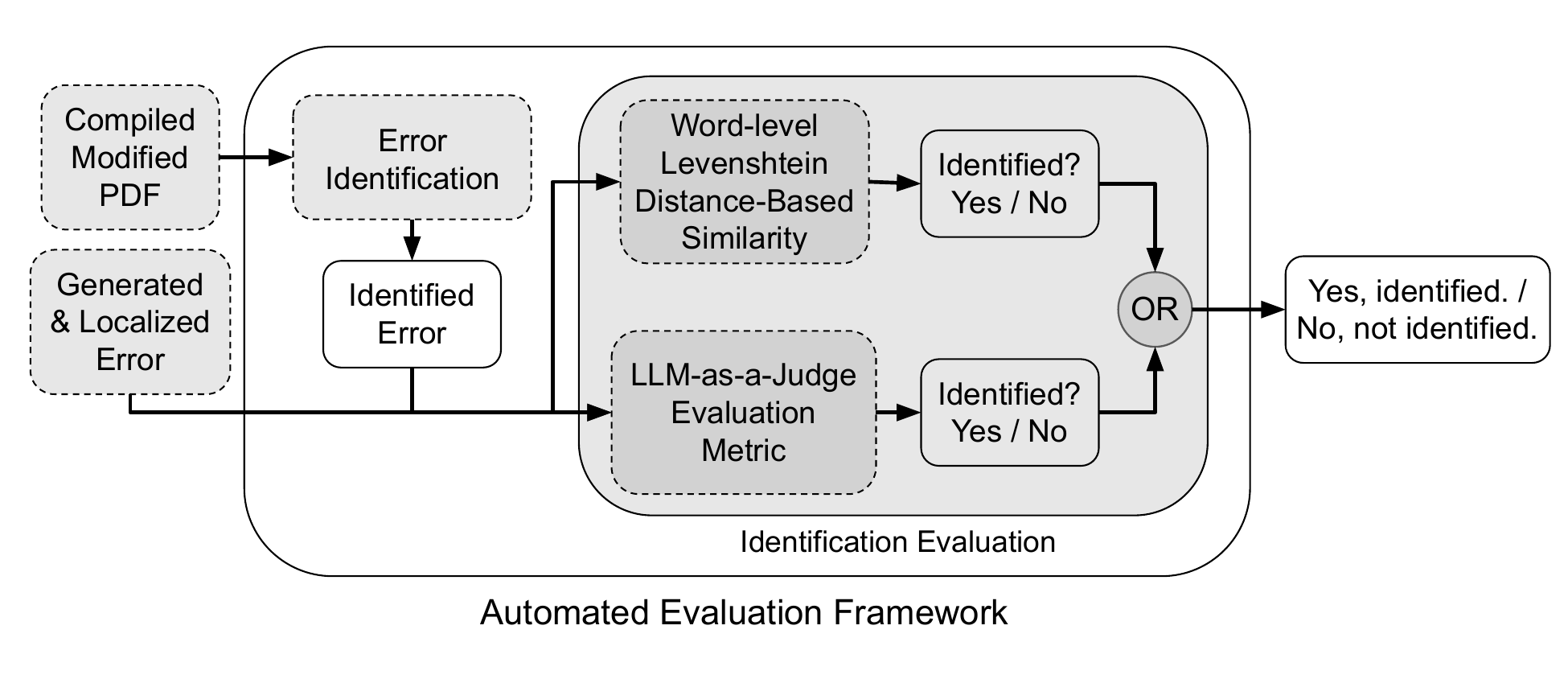}
    \caption{Detailed automated evaluation framework workflow. }
    \label{fig:evaluation_framework}
\end{figure}

With the constructed benchmark, we can now evaluate how well different LLMs are at identifying the introduced synthetic errors. We use an automated evaluation framework to do so. It takes in the outputs from the constructed benchmark --- the compiled modified PDF, the generated errors, and the localized errors --- then performs error identification and identification evaluation. We first discuss the error identification task, and then detail how we evaluate the identified errors, going into the specific evaluation metrics. Figure \ref{fig:evaluation_framework} outlines the workflow of the automated evaluation framework.

\subsection{Error Identification}
\label{subsec:error_id}

\paragraph{LLM error identification and localization.} We evaluate each model by prompting it to identify erroneous text excerpts from the compiled PDF of each paper. Importantly, we prompt the model to specifically identify errors instead of generating a holistic review, which has been shown to be more effective for error identification \citep{liu2023}. The model receives the full PDF and is asked to return excerpts corresponding to any identified errors, ranked in order of significance. This differentiates from the internal identification process, where we prompt the LLM with the LaTeX source. The full prompt is provided in Appendix \ref{appendix:error_identification}. If any returned excerpt overlaps with passages in the ground truth error excerpt set, we consider the error successfully identified. This setup reflects a realistic peer-review scenario, where a human reviewer could locate the error as long as at least one highlighted passage points to a portion of the correct error text.

\paragraph{Constraints to prevent gaming.} To prevent models from trivially inflating their similarity scores, we impose two constraints in the error identification prompt (see Appendix \ref{appendix:error_identification} for details): 

\begin{itemize}
    \item \textit{Maximum number of excerpts}. Each model may return at most 10 excerpts, ranked by importance, with the most significant first.
    \item \textit{Maximum excerpt length}. Each excerpt is restricted to the maximum length of the ground truth error excerpt that pertains to the error.
\end{itemize}

These constraints prevent trivial strategies such as highlighting the entire paper or all sentences sequentially, which could artificially inflate identification scores without demonstrating true error identification ability. For example, returning every sentence or the full paper as a single excerpt would succeed but provide no meaningful evaluation. By limiting both the number and length of excerpts, and requiring prioritization, the evaluation measures meaningful error identification rather than exploitative behavior.

\subsection{Identification Evaluation}
In this section, we introduce the two metrics that are used to evaluate error identification. The first is a word-level Levenshtein distance-based metric that measures the similarity between passages. The second is an LLM-as-a-judge response to evaluate whether an error text excerpt has been identified. We use these two metrics and consider that an error has been successfully identified and localized if it is flagged by either metric. This is a more lenient metric compared to what we used in Section \ref{section:multi_step_filtration} for internal identification, where we only used the word-level Levenshtein distance-based metric. The reliability of this approach is analyzed more in Section \ref{section:manual_annotation}. Preliminary testing on a small set of samples indicated that this approach best aligns with manual error identification (see Appendix \ref{appendix:preliminary_eval_exp} for details). In this section, we provide details on these two metrics.

\subsubsection{Word-Level Levenshtein Distance-Based Similarity}
\label{section:levenshtein_distance}
\paragraph{Motivation for metric choice.} When choosing a distance metric, we considered several options for quantifying similarity between the text returned by the identification model and ground truth error excerpts. Purely semantic similarity measures, such as cosine similarity on embeddings, are insufficient in this context. Papers often contain multiple passages with similar overall meaning, which can produce high semantic similarity scores even if the model has not actually highlighted the specific erroneous text. Conversely, character-level metrics are brittle, over-penalizing minor lexical variations, punctuation differences, or formatting inconsistencies, which are common in LLM-generated text. Word-level comparison provides a balance: it captures meaningful lexical correspondences while remaining robust to superficial formatting changes, and it is computationally more efficient compared to character-level metrics.

Preserving word order is also crucial for comparing the semantic meaning of the text. Token-set measures, such as Jaccard similarity, fail to account for re-orderings that alter semantic meaning (e.g., “algorithm A is better than B” vs.\ “algorithm B is better than A”). For these reasons, we adopt a \textit{word-level Levenshtein distance-based metric}, which measures the minimum number of word edits required to transform one text sequence into another. Levenshtein distance has also been applied to other contexts evaluating LLM outputs, such as code generation \citep{pan2025} or how LLM learn numerical representations \citep{marjieh2025}. In our context, our metric effectively balances lexical precision, robustness to minor variations, and preservation of semantic order, making it well-suited for evaluating whether a model has correctly identified a text excerpt related to the inserted error. 

\paragraph{Metric details.} For each identification model on a paper-error pair, we calculate this metric between each LLM-identified error text excerpt and each ground truth error text excerpt. If one of the identified excerpts is similar enough, as measured by this metric, to one of the ground truth excerpts, the error is considered identified. Now we give a formal definition to this metric used. Let $\mathbb{X}$ denote the set of modified text chunks containing the injected error, and $\mathbb{Y}$ denote the set of excerpts returned by the LLM. For each $x_i \in \mathbb{X}$ and $y_j \in \mathbb{Y}$, we measure the degree of overlap to determine whether the error has been successfully identified. The word-level Levenshtein distance \citep{levenshtein1966} counts the minimum number of word \textit{insertions, deletions,} and \textit{substitutions} required to transform $x_i$ into $y_j$. We convert this into a normalized similarity score:

\[
S_{\mathrm{edit}}(x_i, y_j) = 1 - \frac{\text{Levenshtein-distance}(x_i, y_j)}{\max\{|x_i|, |y_j|\}},
\]
where $S_{\mathrm{edit}} = 1$ indicates an exact match, and values close to 1 indicate minor word-level differences.

To handle cases where the LLM output may correspond to a subset or superset of the ground-truth chunk, we split both $x_i$ and $y_j$ into sentences and compute a symmetric sub-span similarity over the set of contiguous sequences of sentences for each of them, defined as $CSS(x_i)$ and $CSS(y_j)$:

\[
S(x_i, y_j) = \max \Biggl\{
\max_{s \subseteq CSS(x_i)} S_{\mathrm{edit}}(s, y_j),\;
\max_{t \subseteq CSS(y_j)} S_{\mathrm{edit}}(x_i, t)
\Biggr\},
\]
where $s$ and $t$ range over contiguous sentence sub-spans of $x_i$ and $y_j$, respectively. The similarity $S(x_i, y_j)$ represents the maximum normalized word-level similarity between any contiguous sentence sub-span of one text and the entirety of the other. An error is considered successfully identified if $S(x_i, y_j) > 0.5$. Using sentence-level sub-spans is sufficient as each inserted error typically spans a few sentences, and LLMs generally output full sentences when highlighting errors. Therefore, it is unnecessary to enumerate all contiguous word-level sub-spans, which would be computationally expensive and provide little additional benefit. 

\subsubsection{LLM-as-a-Judge Evaluation Metric}
LLM-as-a-judge is commonly used for scoring, ranking, and other ML tasks for evaluation \citep{li2025llm}. In our use case, in addition to the word-level Levenshtein distance-based metric, we prompt an instance of the insertion LLM to decide whether the text excerpts produced by the identification models correspond to the ground-truth error texts. For each identified error excerpt, the LLM will output whether the identification was correct or not. If any of the identified errors corresponded to one of the ground truth error texts, the identification is considered successful overall. The full prompt is provided in Appendix \ref{appendix:error_evaluation}.

\section{Results and Analysis}
\label{section:results}

Using \benchmark{}, we assess the ability of five state-of-the-art LLMs to identify errors in academic text. In this section, we first present results demonstrating the reliability and validity of our evaluation metric, and then report the error identification performance of each model as measured by our framework, providing a ranking of their performance.

\subsection{Reliability of Automated Insertion and Evaluation}
\label{section:manual_annotation}
To validate the reliability of our evaluation metric, we compare its outputs against human annotations. Our validation focuses on two aspects: (i) whether the inserted perturbation constitutes a valid scientific error, and (ii) whether the automated system correctly identifies the inserted error. 

In total, we evaluate 59 \gemini{}-inserted and 59 \gpt{}-inserted errors across the same set of 59 papers. We find that 53/59 (90\%) of the \gemini{}-inserted errors and 57/59 (97\%) of the \gpt{}-inserted errors are judged to be valid scientific errors. In this set, the invalid examples consists of cases where the inserted modification restates assumptions, reframes simplifications of the problem formulation, or highlights existing limitations already implicit in the paper, instead of introducing new errors. We do not observe any surface inconsistencies --- the inserted errors consistently preserve the structure and coherence of the original content. We do not observe any cases where the modified text causes surface-level inconsistency in the paper. To further assess the reliability of the error identification task, we conduct a more detailed human evaluation on a randomly selected subset of 29 papers from this set. For each paper, we collect manual judgments for all five identification models. Annotation is performed by three graduate-level machine learning researchers, with each paper independently labeled by two annotators. We use only cases with agreement between annotators that the inserted error is valid when comparing against automated evaluation outputs.

\paragraph{Inter-annotator agreement.} Across the 29 sampled papers, all 29 \gemini{}-inserted errors were judged valid by both annotators, and 28 of the 29 \gpt{}-inserted errors were judged valid. Annotators showed perfect agreement (100\%) on error validity. For the subset of errors deemed valid, we further collected human annotations on whether each of the five identification models successfully identified the inserted error. Across $28\times5+29\times5=285$ error identifications, we find that there were 32 (11\%) disagreements, leaving 253 cases where the annotators agreed (refer to Appendix \ref{appendix:inter_annotator_agreement} for a more detailed breakdown of agreement). To evaluate reliability more rigorously, we computed Krippendorff's alpha \citep{krippendorff2011}, $\alpha$, which ranges from -1 to 1, where 1 indicates perfect agreement, 0 indicates chance-level agreement, and -1 indicates perfect disagreement. We obtain $\alpha = 0.73$ with a 95\% confidence interval of $[0.62, 0.80]$, indicating substantial inter-annotator agreement. This reliability is consistent with prior studies reporting human agreement on similar annotation tasks \citep{guo2023, dycke2025}.

\paragraph{Automated evaluation metric versus annotator agreement.} 
We assess how well our automated identification framework aligns with human judgment using the 253 errors for which annotators reached agreement. As described in Section \ref{section:evaluation}, an error is considered successfully identified if it is detected by either the word-level Levenshtein distance metric or the LLM-based identification prompt. Table \ref{tab:automate_vs_human_eval} compares the automated evaluation against human annotations, treating the human labels as ground truth. 

\begin{table}[t]
\centering
\begin{subtable}{0.45\textwidth}
\centering
\begin{tabular}{l|ccc}
\hline
\textbf{Label} & \textbf{Precision} & \textbf{Recall} & \textbf{Support} \\
\hline
0 & 0.99 & 0.97 & 190 \\
1 & 0.91 & 0.98 & 63 \\
\hline
\end{tabular}
\caption{Precision and recall.}
\end{subtable}
\hfill
\begin{subtable}{0.50\textwidth}
\centering
\begin{tabular}{c|cc}
\hline
\diagbox{\textbf{Human}}{\textbf{Auto}} & 0 & 1 \\
\hline
 0 & 184 & 6 \\
 1 & 1 & 62 \\
\hline
\end{tabular}
\caption{Confusion matrix.}
\end{subtable}
\caption{Comparison of the automated evaluation metric against human annotators for error identification. 0 indicates it was not identified; 1 indicates the error was identified.}
\label{tab:automate_vs_human_eval}
\end{table}

To quantify overall agreement, we compute Krippendorff's alpha between the automated metric and human annotators, obtaining $\alpha = 0.93$ with a 95\% confidence interval of $[0.89, 0.96]$. This indicates strong agreement, demonstrating that our automated evaluation metric reliably reflects inter-annotator human judgment in identifying errors.

\subsection{LLM Error Identification Results}
\label{section:identification_results}

We evaluated five state-of-the-art LLMs on the \benchmark{} benchmark. Our evaluation procedure aims to understand LLMs' abilities to aid in human peer review. When deploying LLMs to support human reviewing workloads, it is important to display the most important errors first, since humans can feasibly review only a limited number of errors. As mentioned in Subsection \ref{subsec:error_id}, we instruct models to output candidate errors in sorted order, and we report accuracy across different $k$ values, where $k\in[10]$. One can treat $k$ as a proxy for manual effort, where a higher $k$ requires reviewing more candidates, increasing the manual workload.

Table \ref{tab:identification_results} shows each model's identification accuracy for its top-$k$ candidate error outputs. Over the whole benchmark, \gpt{} has the best performance, achieving an identification accuracy of 39.1\% at $k=10$. However, while these raw accuracies provide an initial comparison, they do not account for the difference in sample sizes (448 \gemini{}-inserted errors vs. 265 \gpt{}-inserted errors) and potential differences in the intrinsic difficulty of the inserted errors. In addition, this also does not account for potential bias in identification results if we use the same insertion and identification model.

\begin{table}[t]
\centering
\begin{tabular}{lclcccc}
\hline
\textbf{Insertion} &\textbf{Size} & \textbf{Identification} & \textbf{k=1} & \textbf{k=3} & \textbf{k=6} & \textbf{k=10}\\
\hline
\multirow{5}{*}{Both}&\multirow{5}{*}{713}&\claude{} & 0.052 & 0.126 & 0.201 & 0.215 \\
&&\deepseek{} & 0.059 & 0.163 & 0.283 & 0.352 \\
&&\gemini{} & 0.066& 0.157 & 0.194 & 0.198 \\
&&\gpt{} & 0.090 & 0.192 & 0.300 & 0.391 \\
&&\grok{} & 0.088 & 0.163 & 0.215 & 0.234 \\
\hline
\multirow{5}{*}{\gemini{}} &\multirow{5}{*}{448}& \claude{} & 0.058 & 0.143 & 0.223 & 0.250\\
&& \deepseek{} & 0.067 & 0.199 & 0.315 & 0.393\\
&& \gemini{} & \textcolor{lightgray}{0.076} & \textcolor{lightgray}{0.190} & \textcolor{lightgray}{0.230} & \textcolor{lightgray}{0.234}\\
&& \gpt{} & 0.103 & \textbf{0.219} & \textbf{0.339} & \textbf{0.440}\\
&& \grok{} & \textbf{0.105} & 0.183 & 0.232 & 0.259\\
\hline
\multirow{5}{*}{\gpt{}} & \multirow{5}{*}{265}&\claude{} & 0.041 & 0.098 & 0.151 & 0.155 \\
&& \deepseek{} & 0.045 & 0.102 & \textbf{0.230} & \textbf{0.283} \\
&& \gemini{} & 0.049& 0.106 & 0.132 & 0.136 \\
&& \gpt{} &\textcolor{lightgray}{0.068} & \textcolor{lightgray}{0.147} & \textcolor{lightgray}{0.234} & \textcolor{lightgray}{0.309} \\
&& \grok{} & \textbf{0.060} & \textbf{0.128} & 0.185 & 0.192 \\
\hline
\end{tabular}
\caption{LLM error identification accuracy at top $k$ candidate error text excerpts. Bolded values indicate the top identification model excluding insertion models. The grayed out results indicate using the same insertion and identification LLM.}
\label{tab:identification_results}
\end{table}

In the literature, logistic regression is often applied in imbalanced settings by estimating coefficients via maximum likelihood over all available data. In particular, the Bradley-Terry coefficient \citep{bradley1952} --- computed from a logistic regression model --- is used to compare the performance and uncertainty estimates of pairwise evaluations in LLMs as well as other settings \citep{chiang2024chatbot, frick2025}. We design our logistic regression model with separate parameters for the insertion models and the identification models. By including insertion model parameters, the model explicitly controls for systematic differences in error difficulty between \gemini{} and \gpt{}. On the other hand, the parameters for the identification models better represent performance differences while accounting for varying error insertion sources. We now formally define our logistic regression model. Let $y_k \in \{0,1\}$ indicate whether a given error is correctly identified within the top-$k$ error candidates. Then, let $X_{\text{insert}} \in \mathbb{X}_{in} := \{\gemini{}, \gpt{}\}$ be the model performing error insertion and $X_{\text{identify}} \in \mathbb{X}_{id} :=\{\claude{}, \deepseek{}, \gemini{}, \gpt{}, \grok{}\}$ be the model performing error identification. Then let the logistic regression coefficients be:
\begin{itemize}
    \item $\beta_{j,k} \in \mathbb{R}$: the coefficient parameterizing the insertion model $j$ at top $k$ candidates.
    \item $\gamma_{i,k} \in \mathbb{R}$: the coefficient parameterizing the identification model $i$ at top $k$ candidates.
    \item $\beta_0 \in \mathbb{R}$: a fixed reference value before adding model-specific effects.
\end{itemize}

We model the probability of a successful identification at each $k$ with insertion model $X_{\text{insert}}$ and identification model $X_{\text{identify}}$ using a logistic regression model where $\sigma(x)=\frac{1}{1+\exp(-x)}$ is the logistic sigmoid and $\mathbb{I}$ is the indicator function:

\begin{equation*}
    f(X_{\text{insert}}, X_{\text{identify}})=\beta_0 + \sum_{i\in{\mathbb{X}_{in}}} \gamma_{i,k}\,\mathbb{I}(X_{\text{insert}} = i)
\;+\;
\sum_{j\in{\mathbb{X}_{id}}} \beta_{j,k}\,\mathbb{I}(X_{\text{identify}} = j)
,
\end{equation*}

\begin{equation*}
    \Pr(y_k = 1 |X_{\text{insert}}, X_{\text{identify}})
    =
    \sigma\!\left( f(X_{\text{insert}}, X_{\text{identify}})
    \right).
\end{equation*}

\begin{figure}[t]
    \centering
    \begin{subfigure}{0.32\textwidth}
        \includegraphics[width=1.0\textwidth]{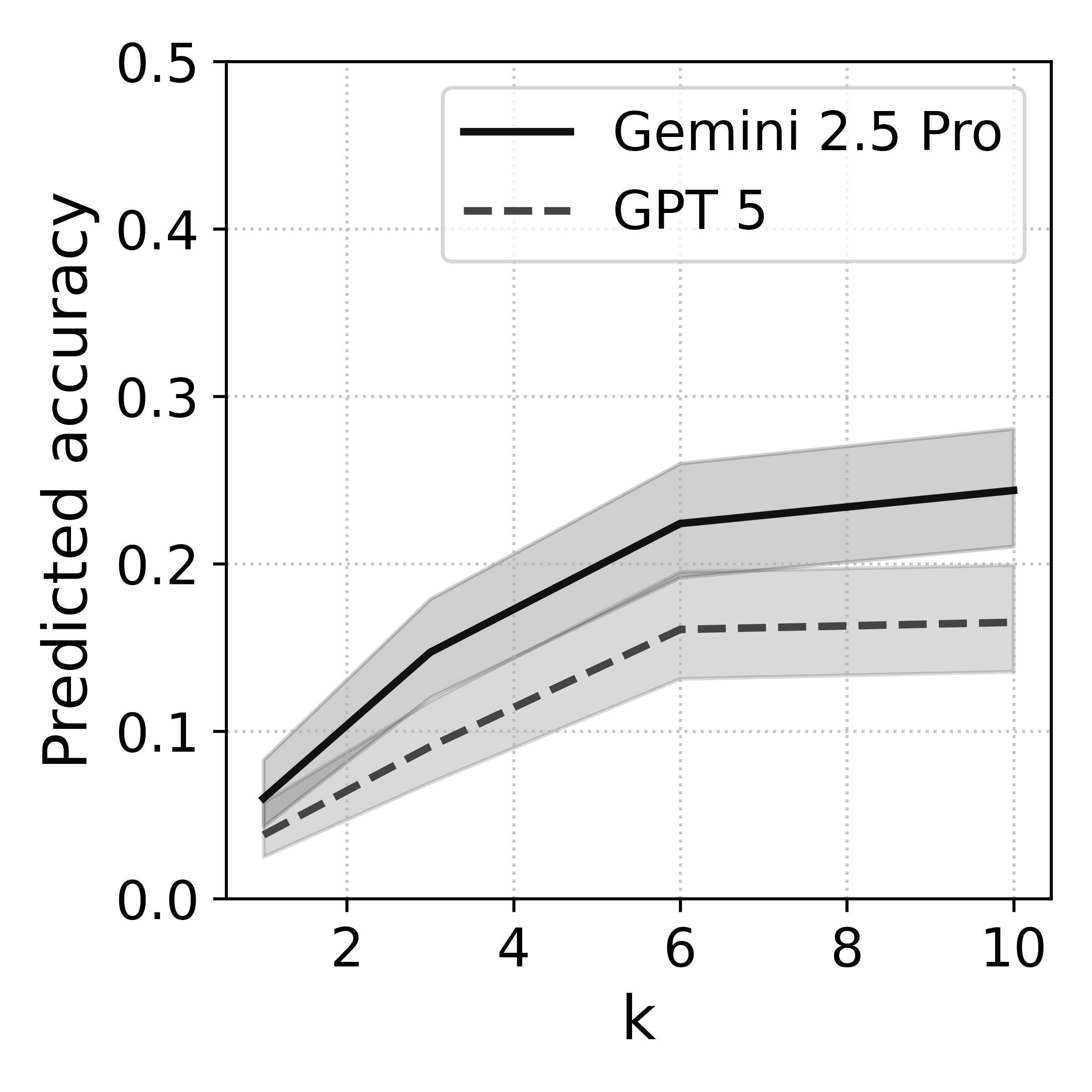}
        \caption{\claude{}}
    \end{subfigure}
    \begin{subfigure}{0.32\textwidth}
        \includegraphics[width=1.0\textwidth]{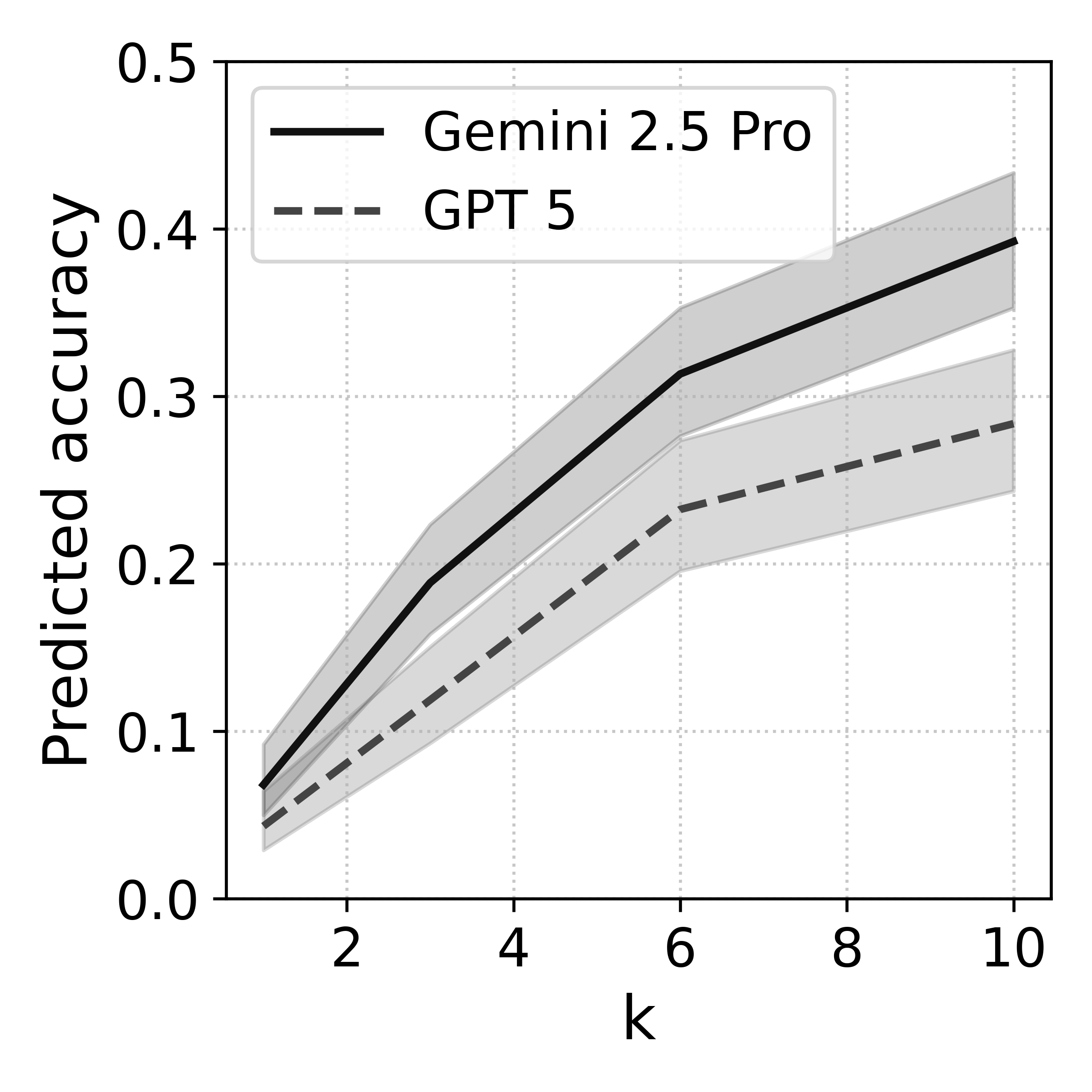}
        \caption{\deepseek{}}
    \end{subfigure}
    \begin{subfigure}{0.32\textwidth}
        \includegraphics[width=1.0\textwidth]{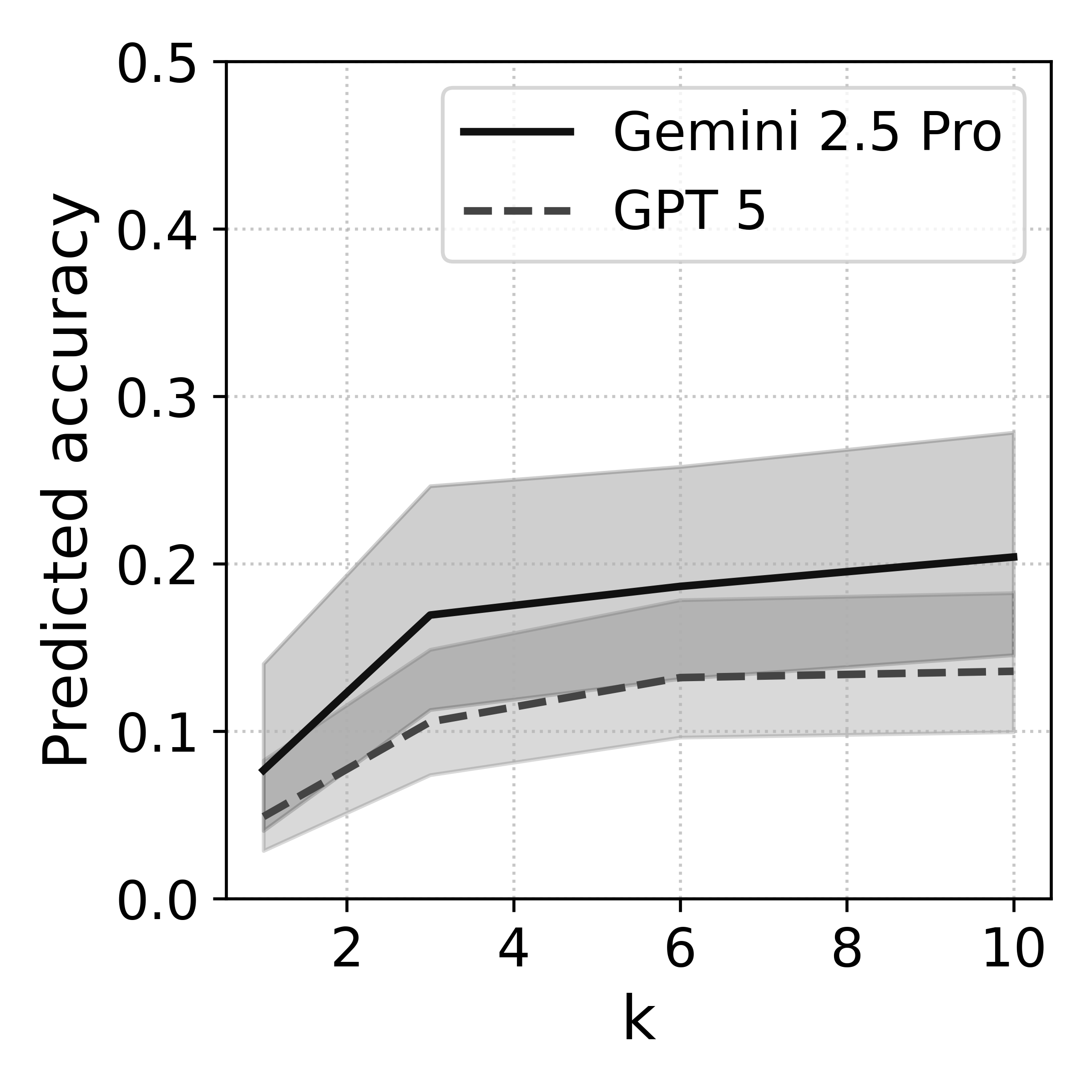}
        \caption{\gemini{}}
    \end{subfigure}
    \begin{subfigure}{0.32\textwidth}
        \includegraphics[width=1.0\textwidth]{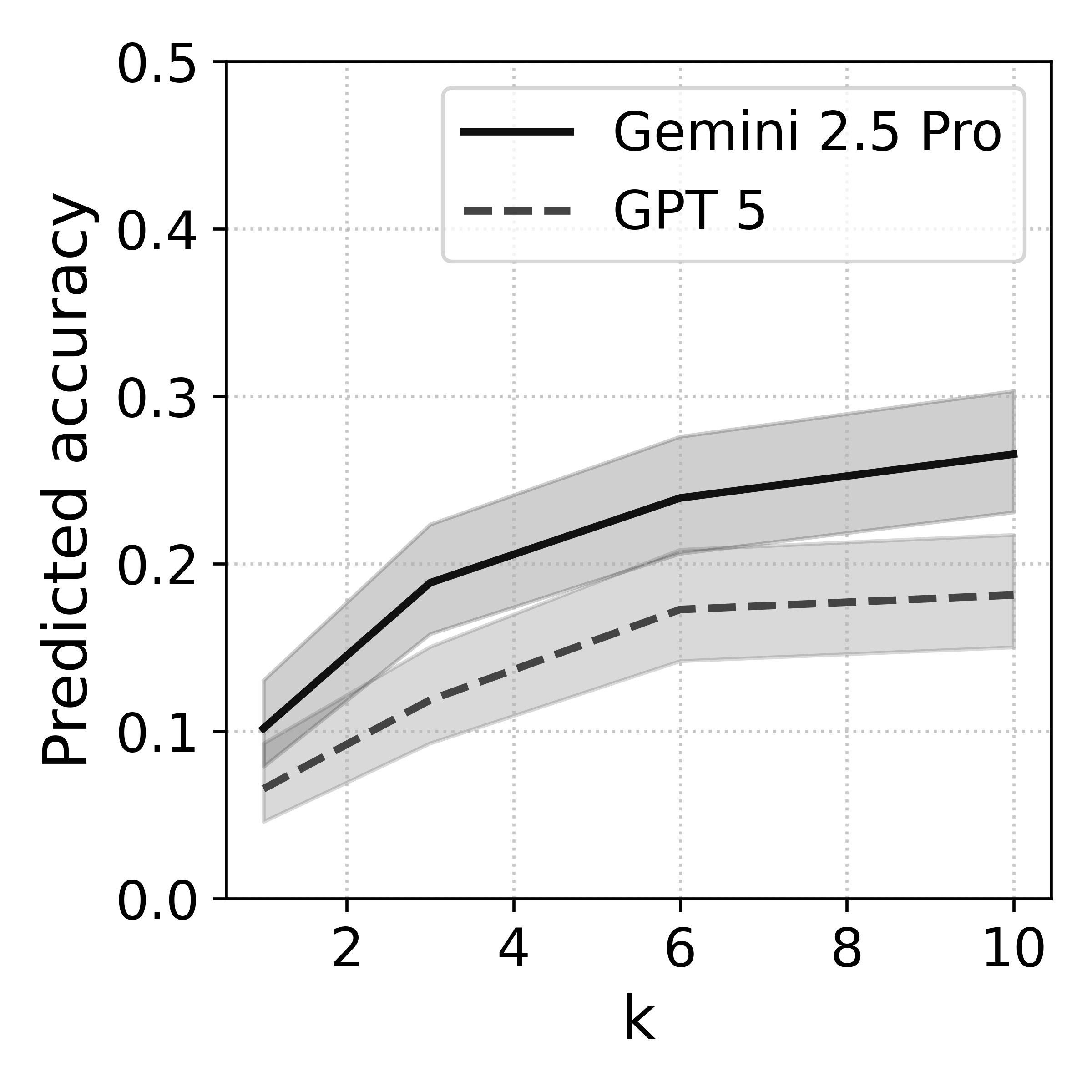}
        \caption{\grok{}}
    \end{subfigure}
    \begin{subfigure}{0.32\textwidth}
        \includegraphics[width=1.0\textwidth]{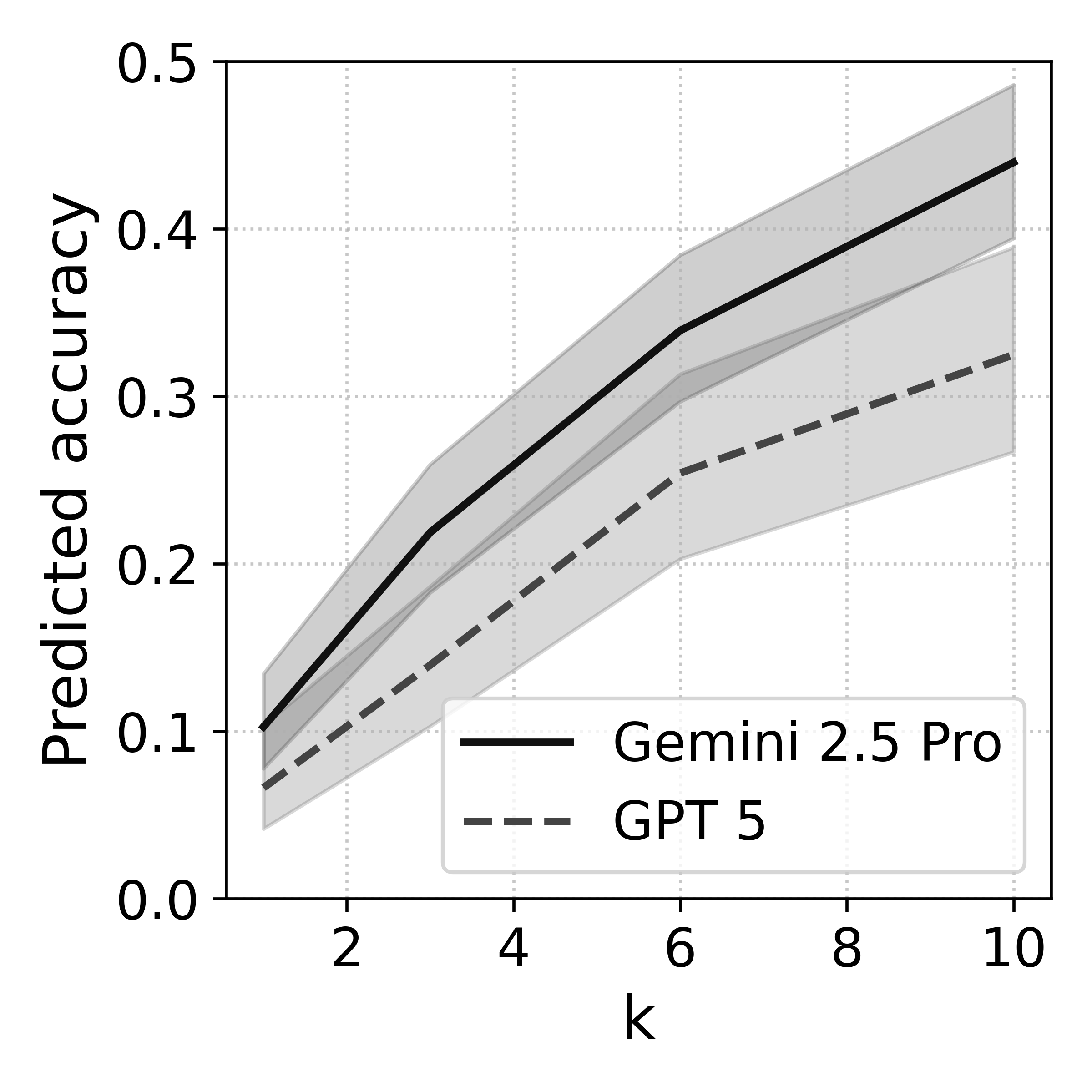}
        \caption{\gpt{}}
    \end{subfigure}
    \caption{ Predicted identification accuracy, $\Pr(y_k=1)$, across top-$k$ error candidates for each identification model. Shaded regions show 95\% confidence intervals.}
    \label{figure:logistic_k_identification}
\end{figure}

To learn the parameters, we fitted the logistic regression on a subset of the identification data. This subset excludes the identification results that used the same insertion and identification LLM to mitigate potential same model biases. This yielded parameter sets of $\beta_{j,k}$ and $\gamma_{i,k}$ that capture how the insertion and identification models, respectively, affect the probability of identifying an error at each $k$. Table \ref{tab:logistic_coefficients} shows the learned coefficients parameterizing the logistic regression model at $k=1,3,6,10$. Furthermore, Figure \ref{figure:logistic_k_identification} shows the resulting identification accuracies, $\Pr(y_k = 1 |X_{\text{insert}}, X_{\text{identify}})$, along with shaded bands representing the model's 95\% confidence interval. We note that the logistic regression model uses the \gemini{} insertion model as a baseline reference and calculates all other coefficients relative to it. In addition, without loss of generality, we use $\beta_0=-3$ (since the model is invariant to constant additions in $f(X_{\text{insert}}, X_{\text{identify}})$) to enforce that all $\beta_{j,k}, \gamma_{i, k} > 0$.

\begin{table}[t]
    \centering
    \begin{tabular}{lcccc}
        \hline
        \textbf{Identification ($j$)} & $\beta_{j,1}$ & $\beta_{j,3}$ & $\beta_{j,6}$ & $\beta_{j,10}$\\
        \hline
        \claude{} & 0.25($\pm$0.18) & 1.24($\pm$0.12) & 1.76($\pm$0.10) & 1.87($\pm$0.10)\\
        \deepseek{} & 0.38($\pm$0.23) & 1.54($\pm$0.15) & 2.22($\pm$0.13) & 2.56($\pm$0.12)\\
        \gemini{} & 0.51($\pm$0.36) & 1.41($\pm$0.25) & 1.53($\pm$0.22) & 1.64($\pm$0.21)\\
        \gpt{} & \textbf{0.83($\pm$0.24)} & \textbf{1.73($\pm$0.17)} & \textbf{2.33($\pm$0.14)} & \textbf{2.76($\pm$0.14)}\\
        \grok{} & 0.82($\pm$0.21) & 1.54($\pm$0.15) & 1.84($\pm$0.13) & 1.98($\pm$0.13)\\
        \hline
        \textbf{Insertion ($i$)} & $\gamma_{i,1}$ & $\gamma_{i,3}$ & $\gamma_{i,6}$ & $\gamma_{i,10}$\\
        \hline
        \gemini{} & 0.0 & 0.0 & 0.0 & 0.0\\
        \gpt{} & -0.48($\pm$0.19) & -0.55($\pm$0.14) & -0.41($\pm$0.11) & -0.49($\pm$0.11)\\
        \hline
    \end{tabular}
    \caption{Learned coefficients and their standard error of the logistic regression model. The coefficients that indicate top error identification ability are bolded. A larger $\beta_{j,k}$ indicates that the model is better at identifying errors. A smaller $\gamma_{i, k}$ indicates that the subset of inserted errors are harder. Note that what determines performance is the \emph{relative} value of the coefficients, not the absolute difference.}
    \label{tab:logistic_coefficients}
\end{table}

\paragraph{Comparison of error identification capabilities.} 

For identification models, we observe that \grok{} and \gpt{} have the highest $\beta_{j,k}$ at $k=1$. This means that they have higher identification accuracies compared to the other models at $k=1$, where the least amount of manual effort is needed. As $k$ increases, \gpt{} remains competitive, but \deepseek{} outperforms \grok{}. To assess overall identification performance across all $k$ values, we compute a single aggregated coefficient for each model: $\beta_j=\frac{1}{10}\sum_{k=1}^{10}\beta_{j,k}$. This approach follows standard practices in information retrieval, where metrics such as Mean Average Precision (MAP) aggregate performance by taking the mean over multiple retrieval depths $k$ \citep{weller2024, wang2024}. We estimate this value and its 95\% confidence interval via a bootstrap procedure with 1000 resamples. For each resample, we sample 713 paper-error pairs with replacement and recompute $\beta_j$. The results are presented in Table \ref{tab:overall_identification_abilities} and Figure \ref{figure:ranking_visualize}, where we confirm that the performance of \gpt{} is overall the best, with the largest $\beta_j$, followed by \deepseek{} and \grok{}. For immediate deployment in peer review workflows, we recommend leveraging \gpt{} as the primary error identification LLM, since it demonstrates the strongest performance overall. We analyze model performance by error type in Appendix \ref{appendix:error_type}, and find that the ranking of models stays relatively consistent across different error types. 

\begin{table}[th!]
\centering
\begin{subtable}[c]{0.5\textwidth}
    \centering
    \includegraphics[width=\linewidth]{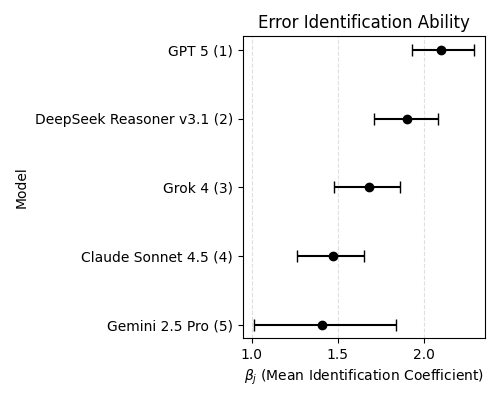}
    \caption{Visualization}
    \label{figure:ranking_visualize}
\end{subtable}
\begin{subtable}[c]{0.55\textwidth}
    \centering
    \begin{tabular}{lccc}
    \hline
    \textbf{Identification Model} & \textbf{$\beta_j$} & \textbf{95\% CI} & \textbf{Rank} \\
    \hline
        \gpt{} & 2.10 & [1.93, 2.29] & 1\\
        \deepseek{} & 1.90 & [1.71, 2.08] & 2\\
        \grok{} & 1.68 & [1.48, 1.86] & 3\\
        \claude{} & 1.47 & [1.26, 1.65] & 4\\
        \gemini{} & 1.41 & [1.01, 1.84] & 5\\
    \hline
    \end{tabular}
    \caption{Summary statistics.}
\end{subtable}
\caption{Overall error identification abilities of LLMs; higher values indicate better performance.}
\label{tab:overall_identification_abilities}
\end{table}

\paragraph{No model surpasses 50\% identification accuracy.} \benchmark{} is designed to be a challenging benchmark. As seen in Table~\ref{tab:identification_results}, no model surpasses 50\% identification accuracy even when considering the top 10 error candidates. In fact, the two worst performing models, \claude{} and \gemini{}, have less than 30\% identification accuracy at $k=10$. At $k=1$ values, where the least manual labor is needed, the identification accuracy is only 5-10\% across all evaluated models, with \gpt{} and \grok{} showing the strongest performance. 

\paragraph{Comparison of the number of error text candidates identified.} At $k=10$, \deepseek{} and \gpt{} achieve the strongest overall performance; however, their accuracy declines more sharply as $k$ decreases. In contrast, \grok{}, \claude{}, and \gemini{} exhibit considerably earlier plateaus. For \gemini{}, accuracy stabilizes almost completely by $k = 4$, while \grok{} and \claude{} flatten around $k = 6$. These differences align with the mean number of error text candidates that each model produces, shown in Table \ref{tab:mean_num_identification}. Although all models are prompted to identify and rank the top 10 most significant errors, they do not necessarily output 10 errors if they cannot detect that many. This is especially evident in the median counts: \deepseek{} and \gpt{} typically produce 9 error spans, whereas \claude{} and \grok{} produce around 5, and \gemini{} produces only 4. These observations have direct implications for how one should choose a model under different manual reviewing capacities. When sufficient manual review capacity is available, \gpt{} and \deepseek{} may be preferable, as their performance scales with larger $k$. In such settings, increasing $k$ beyond 10 may yield even higher identification accuracy. When manual effort is limited, \grok{} is preferred over \deepseek{}. 

\begin{table}[!t]
\centering
\begin{tabular}{llccc}
\hline
\multirow{2}{*}{\textbf{Insertion Model}} & \multirow{2}{*}{\textbf{Identification Model}} & \multicolumn{3}{c}{\textbf{\# Error Texts Candidates}} \\
& & Median & Mean & Std \\
\hline
\multirow{5}{*}{\gemini{}} & \claude{} & 5 & 5.41 & 1.48\\
& \deepseek{} & 9 & 7.77 & 2.78 \\
& \textit{\gemini{}} (insertion model) & 4 & 3.83 & 1.42\\
& \gpt{} & 9 & 8.56 & 1.62\\
& \grok{} & 5 & 4.88 & 2.11\\
\hline
\multirow{5}{*}{\gpt{}} & \claude{} & 5 & 5.32 & 1.56 \\
& \deepseek{} & 9 & 7.65 & 2.98 \\
& \gemini{} & 4 & 3.79 & 1.52 \\
& \textit{\gpt{}}(insertion model) & 9 & 8.43 & 1.71 \\
& \grok{} & 4 & 4.60 & 2.26\\
\hline
\end{tabular}
\caption{Summary for number of error text candidates identified by each model.}
\label{tab:mean_num_identification}
\end{table}

\section{Limitations and Future Work}
In this work, we introduce \benchmark{}, a scalable and automated benchmark consisting of 713 paper-error pairs to effectively evaluate the error identification and localization capabilities of state-of-the-art LLMs, constructed systematically using LLMs themselves. This approach allows us to ensure that the inserted errors are non-trivial, relevant, and explicit, as well as autonomously evaluate LLMs in a scalable manner. Current frontier LLMs have an accuracy of no more than 39.1\% in identifying the errors presented in \benchmark{} when allowed to output 10 error text candidates. In what follows, we discuss limitations and future directions to build on this work. 

While our insertion framework is designed to ensure that inserted errors are relevant and challenging, there is no guarantee that all errors are fully valid. In our manual annotations (Section \ref{section:manual_annotation}), we found that 1 out of 58 errors (1.7\%) was invalid, as it could not be directly identified from the contents of the paper alone, a criterion explicitly imposed during error generation. This underscores a trade-off between the scale of error generation and the extent to which the generated errors meet all validity criteria. In contrast, some related works opt for smaller-scale datasets, manually annotating all papers to better ensure that every error meets predefined criteria \citep{liu2023, son2025}. While this approach generally increases alignment to the intended criteria, it is significantly less scalable compared to automated approaches.

These considerations about error validity and scalability also have implications for the benchmark as a whole. Currently, \benchmark{} is static and evaluates the performance of current frontier LLMs. Over time, it may become outdated as newer generations of LLMs are released and the risk of data contamination increases when models are trained on more recent data \citep{deng2024, sainz2023}. While it is generally challenging to consistently keep augmenting manually-constructed datasets, \benchmark{} is automatically curated, which allows it to be augmented with new peer-reviewed papers to mitigate these issues. However, this process remains resource-intensive and requires extensive filtering: only about 5\% of \gpt{}-generated errors survive the multi-step filtering process (see Section \ref{section:multi_step_filtration} for details). Nevertheless, with sufficient resources, the current automated framework could support the construction of a dynamic benchmark and leaderboard, enabling continuous updates and systematic horizontal evaluation of newer papers and models.

Building on the idea of dynamically updating \benchmark{}, future work could extend the framework to peer-reviewed papers from a broader range of domains. Currently, the main \benchmark{} dataset consists only of papers from ICML 2025, which are primarily focused on AI and machine learning. While we provide an initial assessment of the framework’s application on other domains by running it on 100 non-ML computer science papers (as outlined in Appendix \ref{appendix:extension}), we encourage further expansion beyond computer science. Extending to fields such as physics, economics, and other disciplines, along with increasing the sample size, would enable a more systematic investigation of potential differences in error insertion and LLM-based error identification across diverse types of content.

In addition to exploring different domains, future work can also systematically examine how error insertion varies across different LLMs to investigate and better understand underlying reasons for differing identification performance across models. As seen in Section \ref{section:identification_results}, \gpt{}-generated errors are generally harder to identify than \gemini{}-inserted errors. This trend is also seen in Figure \ref{figure:logistic_k_identification}, where identification accuracy for \gemini{}-inserted errors is consistently higher. These differences may result from either a genuine systematic difference in the error generated by \gemini{} and \gpt{}, or \gemini{} filtering out fewer easy errors than \gpt{} during the multi-step filtration during insertion. Studying this phenomenon systematically could provide insights into both model-specific biases in error generation and error difficulty. Moreover, if \benchmark{} is continuously updated with newer frontier LLMs, it would be valuable to investigate whether the benchmark becomes progressively harder over time or eventually saturates, depending on the capabilities of the models used for both error insertion and identification.

Finally, \benchmark{} provides a testbed for future work to investigate the underlying reasons why LLMs struggle to identify certain errors. Researchers could systematically probe the reasoning processes behind error identification and localization, analyzing how models interpret claims, evidence, and inconsistencies within a paper. Such studies could uncover specific failure modes that are not explicitly addressed in our current work. By better understanding these failure modes, future research can guide the development of more robust LLMs and improve error identification strategies. Ultimately, we hope that \benchmark{} accelerates progress toward LLMs that can help us improve the rigor and reliability of scientific research.

\section*{Acknowledgment}
We gratefully acknowledge the contributions and insights of Alexander Goldberg, Ziming Luo, Nicholas Perello, Vignesh Viswanathan and Anthony Zhou, who helped test and provided feedback on early iterations of the error insertion pipeline. This work was supported by grants NSF CAREER 1942124, NSF 2200410, and ONR N000142512346.

\vskip 0.2in
\bibliography{main}
\bibliographystyle{bib}

\appendix
\section{Error Insertion and Identification Example}
\label{appendix:example_error}
In this section, we detail an example paper-error pair that has passed through all filtering steps during the insertion framework, and also provide the identification outputs.
\paragraph{Paper:} 
\href{https://arxiv.org/abs/2402.07860}{On the Detection of Reviewer-Author Collusion Rings From Paper Bidding}
\subsection{Insertion Outputs}

\begin{tcolorbox}
[breakable,colback=black!5!white,colframe=black!20!white,title=\textbf{\textcolor{black}{Claim Extracted using \gemini}}]
Groups of colluding reviewers can successfully manipulate the paper assignment process to achieve assignment to a substantial fraction of their members' papers (up to 30\% in the AAMAS dataset and 24\% in the S2ORC dataset) while using a bidding strategy that remains undetected by common dense-subgraph discovery algorithms.
\end{tcolorbox}

\begin{tcolorbox}
[breakable,colback=black!5!white,colframe=black!20!white,title=\textbf{\textcolor{black}{Error Generated using \gemini}}]

:original text:\\
In our experiments, we use \lstinline|$\simfn(\arbrev, \arbpap) = \frac{1}{2} \txtsim(\arbrev, \arbpap) 2^{\mathbb{I}[(\arbrev, \arbpap) \in \bidset]}$| based on the function used in the 2016 Conference on Neural Information Processing Systems (NeurIPS), a large machine-learning conference$\sim\backslash \text{citep}\{\text{shah2018design}\}$.\\

:modified text:\\
In our experiments, we use \lstinline|$\simfn(\arbrev, \arbpap) = \frac{1}{2} \txtsim(\arbrev, \arbpap) + 0.1 \cdot \mathbb{I}[(\arbrev, \arbpap) \in \bidset]$| based on the function used in the 2016 Conference on Neural Information Processing Systems (NeurIPS), a large machine-learning conference$\sim\backslash \text{citep}\{\text{shah2018design}\}$.\\

:explanation:\\
The conceptual error introduced is a subtle but critical change to the similarity function that determines the paper assignment. This function is central to evaluating the ``success'' of a collusion ring's manipulation.\\

1.  **Original Model**: The original similarity function, \lstinline|`$\simfn(\arbrev, \arbpap) = \frac{1}{2} \txtsim(\arbrev, \arbpap) 2^{\mathbb{I}[(\arbrev, \arbpap) \in \bidset]}$'|, uses a multiplicative bonus for a positive bid. A bid \lstinline|(`$\mathbb{I}=1$')| doubles the total similarity score, making bids extremely powerful in the maximum-similarity assignment process. This is what allows a colluder's strategic bid to have a high probability of resulting in a desired assignment.\\

2.  **Modified Model**: The modified function, \lstinline|`$\simfn(\arbrev, \arbpap) = \frac{1}{2} \txtsim(\arbrev, \arbpap) + 0.1 \cdot \mathbb{I}[(\arbrev, \arbpap) \in \bidset]$'|, changes the bid's effect from a powerful multiplicative bonus to a small additive constant \lstinline|(`+0.1')|.\\

3.  **The Flaw**: This change dramatically weakens the influence of bidding on the final paper assignment. The text-similarity scores \lstinline|`$\txtsim(\arbrev, \arbpap)$'| are values in \lstinline|`$[0, 1]$'|, and can be relatively high (mean of 0.52 in the S2ORC dataset). An additive bonus of 0.1 provides only a minor boost to the similarity score, which is often insufficient to override assignments that would have been made based on high text similarity alone. Consequently, the colluders' strategic bidding would be far less effective than claimed. The paper's conclusion that colluders can achieve assignment to a ``substantial fraction" of their papers (up to 30\%) is no longer supported by the described methodology. The core claim about collusion being both successful and undetectable is thus undermined on the ``success" dimension.\\

This error is non-trivial because the formula appears plausible at a glance, and detecting the flaw requires a deeper understanding of how max-weight matching assignment algorithms work and a consideration of the typical magnitude of the text-similarity scores.
\end{tcolorbox}

\begin{tcolorbox}
[breakable,colback=black!5!white,colframe=black!20!white,title=\textbf{\textcolor{black}{Error Localized for Insertion using \gemini}}]
Error in assumptions (core conceptual)
error:\\
We find that collusion rings can achieve considerable success at manipulating the paper assignment while remaining hidden from detection: for example, in one dataset, undetected colluders are able to achieve assignment to up to 30\% of the papers authored by other colluders.\\

error:\\
$\backslash$item Colluders are able to achieve assignment to a substantial fraction of the papers authored by other colluders while avoiding detection by all algorithms (30\% and 24\% in each of the two datasets).\\

error:\\
$\backslash$item A sizeable fraction of colluders can get at least one of their papers reviewed by another colluder while avoiding detection (54\% and 35\% in each of the two datasets).\\ 

error:\\
On AAMAS, we see that at \lstinline|$(\size=16, \densitybp=0.8)$| where no detection algorithm was able to detect colluders with high probability, the colluders can successfully achieve assignments to $30\%$ of their target papers and $54\%$ of the colluders.\\

error:\\
Still, in the cell \lstinline|$(\size=26, \densitybp=0.8)$| where the detection algorithms performed poorly, colluders can achieve assignment to $26\%$  of target papers and $42\%$ of colluders, a sizeable influence on the paper assignment.
\end{tcolorbox}
\subsection{Identification Outputs}
As a reminder, we consider an error successfully identified if one of the located error excerpts (generated error excerpts $\cup$ localized error excerpts) match with the ground truth error excerpts by either using the world-level Levenshtein distance-based similarity or LLM-as-a-judge. LLM-as-a-judge = 0 indicates incorrect error excerpt; LLM-as-a-judge = 1 indicates correct error excerpt.

\begin{tcolorbox} [breakable,colback=black!5!white,colframe=black!20!white,title=\textbf{\textcolor{black}{Failed Identification using \gemini}}]
1 :error text: \underline{(Levenshtein based similarity $<$ 0.5, LLM-as-a-judge = 0)}\\ 
Since the dataset does not include authorship information, we reconstruct the authorships A by subsampling 3 conflicts-of-interest uniformly at random for each paper. The resulting dataset has 526 papers and 596 reviewers, 398 of whom authored at least one paper. The dataset also does not contain text-similarity scores T(r,p). We generate synthetic text-similarity scores using the procedure described in Appendix A, based on the text-similarities from the 2018 International Conference on Learning Representations reconstructed by Xu et al. (2019).\\

2 :error text: \underline{(Levenshtein based similarity $<$ 0.5, LLM-as-a-judge = 0)}\\
Since this method operates on undirected graphs and is non-trivial to adapt to a directed setting, we first map our bidding graph G1 to an undirected graph G' = (V1, E') before inputting it to this algorithm. The input graph G' has the same vertex set as G1 and has an edge (r1,r2) iff both edges (r1,r2) $\in$ E1 and (r2,r1) $\in$ E1; that is, E' = {(r1,r2) $\in$ R² : (r1, r2) $\in$ E1$\land$(r2, r1) $\in$ E1}).\\

3 :error text: \underline{(Levenshtein based similarity $<$ 0.5, LLM-as-a-judge = 0)}\\
We find that the detection performance of OQC-Local is significantly better when only the heuristic initialization is used, since the random initializations often resulted in output with a higher objective value but lower overlap with the true colluders; thus, we show the results with heuristic initialization only in this section and defer the results with all initializations to Appendix B. However, the poor performance of OQC-Local when all initializations are used indicates that the objective value of OQC-Local is misaligned with the detection objective.\\

4 :error text: \underline{(Levenshtein based similarity $<$ 0.5, LLM-as-a-judge = 0)}\\
For each setting of (k, $\gamma$), we choose a subset of k reviewers uniformly at random from among those reviewers that authored at least one paper. We then add edges uniformly at random between reviewers until the subgraph has edge density at least $\gamma$. This modified graph is then passed as input to each detection algorithm. We repeat this procedure for 50 trials for each setting.\\
\end{tcolorbox}
\begin{tcolorbox}[breakable,colback=black!5!white,colframe=black!20!white,title=\textbf{\textcolor{black}{Succeeded Identification using \grok}}]
1 :error text: \underline{(Levenshtein based similarity $<$ 0.5, LLM-as-a-judge = 0)}\\
Since the dataset does not include authorship information, we reconstruct the authorships A by subsampling 3 conflicts-of-interest uniformly at random for each paper. The resulting dataset has 526 papers and 596 reviewers, 398 of whom authored at least one paper. The dataset also does not contain text-similarity scores T(r, p). We generate synthetic text-similarity scores using the procedure described in Appendix A, based on the text-similarities from the 2018 International Conference on Learning Representations reconstructed by Xu et al. (2019).\\

2 :error text: \underline{(Levenshtein based similarity $<$ 0.5, LLM-as-a-judge = 0)}\\
Our second dataset, which we refer to as ``S2ORC", is the semi-synthetic dataset constructed and made publicly available by Wu et al. (2021). This dataset contains synthetic bids between a large subset of published computer science papers and authors from the Semantic Scholar Open Research Corpus (Ammar
et al., 2018), designed to match statistics from the NeurIPS 2016 conference (Shah et al., 2018).\\

3 :error text: \underline{(Levenshtein based similarity $<$ 0.5, LLM-as-a-judge = 0)}\\
We suppose that there exists a group of colluding reviewers $M \subset R$ who try to manipulate the paper assignment by altering their bids, with the aim of being assigned to review the papers authored by other members of the colluding group. The objective of a collusion-detection algorithm is to output M given the set of bids B, along with the authorships A and the other conflicts C. Note that the text-similarities T(r, p) cannot be used for detection in our analysis–we consider the problem of detection using only the bidding itself and the authorships.\\

4 :error text: \underline{(Levenshtein based similarity $<$ 0.5, LLM-as-a-judge = 0)}\\
As our original datasets do not contain colluding reviewers, we inject collusion into the datasets by choosing a group of reviewers to be the colluders M and modifying the bids of these reviewers (i.e., adding or removing elements of M×P to/from B). The strongest form of collusion would be to add bids between all reviewers in M and all papers authored by other reviewers in M, while additionally removing all other bids by reviewers in M.\\

5 :error text: \underline{(Levenshtein based similarity $<$ 0.5, LLM-as-a-judge = 0)}\\
We consider a simplified setting with binary bids (positive or neutral); note that allowing additional levels of bids can only give colluders more flexibility to manipulate the paper assignment.\\

6 :error text: \underline{(\textbf{Levenshtein based similarity = 0.62 $>$ 0.5}, LLM-as-a-judge = 0)}\\
In our experiments, we use S(r, p) = 1 2T(r, p)+0.1·I[(r, p) $\in$ B] based on the function used in the 2016 Conference on Neural Information Processing Systems (NeurIPS), a large machine-learning conference (Shah et al., 2018).\\

7 :error text: \underline{(Levenshtein based similarity $<$ 0.5, LLM-as-a-judge = 0)}\\
we assume that both of these datasets contain only bids from ``honest” (non-colluding) reviewers. The AAMAS dataset contains information from reviewers that did not opt-out from the data collection, and we expect that any colluding reviewers in the conference would have done so. The S2ORC bids are synthetic and do not model malicious reviewer behavior.
\end{tcolorbox}
\section{LLM Prompts Used}
\subsection{Claim Extraction}
\begin{tcolorbox}[breakable, colback=black!5!white,colframe=black!20!white,title=\textbf{\textcolor{black}{Claim Extraction Prompt}}]\label{appendix:claim_extraction}
\small
    I will provide you with the full LaTeX source of a research paper. Your task is to read the paper thoroughly and list out all major, unique, and falsifiable claims made in the paper. The claims should follow these rules:\\

    1. Each claim pertains to the core scientific content of the paper and its contributions.
    
    2. Each claim is unique.
    
    3. You should not have repetitive claims or claims that are too similar to each other.\\
    
    Please number the claims and return the output in this exact format:\\
    
    1. $<$description of first falsifiable claim$>$
    
    2. $<$description of second falsifiable claim$>$
    
    3. ...\\

    (...there can be as many falsifiable claims as necessary...)

\end{tcolorbox}

\subsection{Error Generation}
\begin{tcolorbox}[breakable, colback=black!5!white,colframe=black!20!white,title=\textbf{\textcolor{black}{Error Generation Prompt}}]\label{appendix:error_generate}
\small
    I am developing a research benchmark purely for research purposes to evaluate how effectively language models (LLMs) can detect errors in academic papers. Your task is to modify sections of a research paper in a way that introduces a non-trivial, plausible conceptual or theoretical oversight, grounded in domain-specific knowledge.\\

    This benchmark aims to test LLMs' capacity for deep understanding, contextual inference, and expert-level critique, rather than surface-level textual correction.\\
    
    Task Instructions\\\\
    You will be given:\\
    1. The input LaTeX source of a paper.
    
    2. A key claim made in the paper.\\
    
    Your job is to introduce a sophisticated, realistic academic error that undermines or weakens the provided claim. Try to introduce an error that would be a core conceptual error. This error should be coherent throughout the text and should:\\
    
    1. Be subtle but significant, and potentially span multiple sections if needed for coherence.
    
    2. Avoid obvious mistakes (e.g., grammar, formatting, numerical typos, misused citations).
    
    3. Be non-trivial to detect without domain expertise, ideally requiring graduate-level reasoning or deeper.
    
    4. Be internally consistent and plausible-the modified paper should still appear well-reasoned and coherent to a casual or intermediate reader.
    
    5. Not be simple or obvious where one sentence is the direct contradiction of it's neighboring sentences. 
    
    6. Be similar to a GENUINE, REALISTIC mistake made by the authors of the paper.
    
    7. Not be easily identified by a master student who has not read the paper in depth.\\
    
    - Make sure that the error is identifiable through the text alone. 
    
    - DO NOT state any inconsistencies or limitations of the original input latex source.
    
    - DO NOT correct any mistakes in the paper. Your job is to INTRODUCE an error NOT correct an error.\\
    
    Modification Format\\
    For each change, return:\\
    
    :original-text:\\
    $<$EXACT LaTeX excerpt from the original source --- copy-pasted with NO modifications$>$\\
    
    :modified-text:\\
    $<$Modified excerpt that is LaTeX compilable --- almost identical to the original but with a plausible conceptual error inserted$>$\\
    
    Repeat original/modified pairs as needed if the error spans multiple non-contiguous parts.\\
    
    STRICT FORMATTING RULES for the LaTeX source original-text:\\
    
    1. Use PRECISE copy and paste methods from the LaTeX input to confirm excerpts are completely identical to the original source. 
    
    2. DO NOT ALTER OR OMIT any token, including $\$$, $\{$, $\}$, or spacing.
    
    3. DO NOT use quotation marks around the error excerpts.
    
    4. DO NOT use ellipses (`...') within a single excerpt. If the original error texts are non-consecutive, return them as separate excerpts.
    
    5. DO NOT omit any $\backslash \text{emph}\{\}$, $\backslash \text{textbf}\{\}$, $\backslash \text{textit}\{\}$ or other formatting that is in the original LaTeX text. 
    
    6. DO NOT hallucinate additional text that is not in the LaTeX source. 
    
    7. DO NOT ALTER OR OMIT any line breaks or new lines.
    
    8. DO NOT interchange punctuations like `` and " etc.
    
    9. The excerpt you output must be exactly matchable in the input latex source!\\
    
    :explanation:\\
    $<$ A brief but clear explanation of what conceptual error was introduced and why it is wrong, especially from a theoretical or domain-specific perspective.$>$
\end{tcolorbox}

\subsection{Invalid Error Filtering}
\begin{tcolorbox}[breakable, colback=black!5!white,colframe=black!20!white,title=\textbf{\textcolor{black}{Invalid Error Filtering Prompt}}]\label{appendix:invalid_error_filter}
\small
    I am developing a research benchmark to evaluate how effectively language models (LLMs) can detect errors in academic papers. I have modified a particular section in the given paper in a way that introduces a non-trivial, plausible conceptual or theoretical oversight, grounded in domain-specific knowledge.\\

    To modify the section, I have taken a claim in the paper, and introduced a sophisticated, realistic academic error that undermines or weakens the provided claim. \\
    
    I am going to provide you with the following:\\
    
    1. Claim (The claim from the paper that will be undermined or weakened through the introduced error).
    
    2. Original text (Part of the original latex source that was altered, i.e. such that error is introduced).
    
    3. Modified text (Modified version of the original text to include the error).
    
    4. Explanation (The explanation behind the introduced error).
    
    5. Input latex source of the paper.\\
    
    Your job is very simple. You need to ensure that the error in the paper I have introduced through the modified text satisfies the below two sanity checks.\\
    
    (a) The Modified Text should not explicitly state that there is an error.
    \\
    (b) The Modified Text should not just highlight an existing weakness, shortcoming, or error in the paper.
    \\
    (c) The Modified Text should actually INTRODUCE a non-trivial, plausible conceptual or theoretical oversight, grounded in domain-specific knowledge.
    \\

    If the Modified Text satisfies the above three criteria, then simply output ``No changes required" without any other padding text.\\

    If the Modified Text does not satisfy all the above criteria or does not contain an error, then simply output ``Filtering required" without any other padding text.
\end{tcolorbox}

\subsection{Easy Error Filtering}
\begin{tcolorbox}[breakable, colback=black!5!white,colframe=black!20!white,title=\textbf{\textcolor{black}{Easy Error Filtering Prompt}}]\label{appendix:easy_error_filter}
\small
    I am developing a research benchmark to evaluate how effectively language models (LLMs) can detect errors in academic papers. I have modified a particular section in the given paper in a way that introduces a non-trivial, plausible conceptual or theoretical oversight, grounded in domain-specific knowledge.\\

    To modify the section, I have taken a claim in the paper, and introduced a sophisticated, realistic academic error that undermines or weakens the provided claim. \\
    
    I am going to provide you with the following:\\
    
    1. Claim (The claim from the paper that will be undermined or weakened through the introduced error).
    
    2. Original text (Part of the original latex source that was altered, i.e. such that error is introduced).
    
    3. Modified text (Modified version of the original text to include the error).
    
    4. Explanation (The explanation behind the introduced error).
    
    5. Input latex source of the paper.\\
    
    Your job is very simple. You need to ensure that the error in the paper I have introduced through the modified text cannot be easily identified by an LLM if it is just fed the PDF of this paper and asked to review it. By ``cannot be easily identified", I mean it should be identified on average less than around 20\% of the time.\\
    
    If the error introduced IS easily identifiable, then simply output ``Filtering required" without any other padding text.\\
    
    If the error introduced is difficult to identify, then simply output ``No changes required" without any other padding text.
\end{tcolorbox}

\subsection{Error Localization}
\begin{tcolorbox}[breakable, colback=black!5!white,colframe=black!20!white,title=\textbf{\textcolor{black}{Error Localization Prompt}}]\label{appendix:error_location}
\small
    I am developing a research benchmark to evaluate how effectively language models (LLMs) can detect errors in academic papers. I have modified a particular section in the given paper in a way that introduces a non-trivial, plausible conceptual or theoretical oversight, grounded in domain-specific knowledge.\\

    To modify the section, I have taken a claim in the paper, and introduced a sophisticated, realistic academic error that undermines or weakens the provided claim. \\
    
    I am going to provide you with the following:\\
    
    1. Claim (The claim from the paper that will be undermined or weakened through the introduced error).
    
    2. Original text (Part of the original latex source that was altered, i.e. such that error is introduced).
    
    3. Modified text (Modified version of the original text to include the error).
    
    4. Explanation (The explanation behind the introduced error).
    
    5. Input latex source of the paper.\\
    
    Your job is to analyze the way the provided section has been modified and then output the part of the input latex source that is incorrect (contains an error) as a result of this modification. It may be natural to think that the "original text" itself is the part of the paper that contains the error. While this may be the case, it may also happen that the original text in itself is correct but causes one or more other parts of the paper (input latex source) to be incorrect (contains an error). The explanation just explains what the introduced error is. It is NOT the part of the input latex source that contains the error.\\
    
    Output ONLY the EXACT LaTeX excerpt from the input latex source that is incorrect — copy-pasted with NO modifications, and do not output any padding text. START printing the LaTeX excerpts by printing ``error:" followed by the incorrect excerpt(s).\\
    
    STRICT FORMATTING RULES for the LaTeX source original-text:\\
    
    1. Use PRECISE copy and paste methods from the LaTeX input to confirm excerpts are completely identical to the original source. 
    
    2. DO NOT ALTER OR OMIT any token, including $\$$, $\{$, $\}$, or spacing.
    
    3. DO NOT use quotation marks around the error excerpts.
    
    4. DO NOT use ellipses (`...') within a single excerpt. If the original error texts are non-consecutive, return them as separate excerpts.
    
    5. DO NOT omit any $\backslash \text{emph}\{\}$, $\backslash \text{textbf}\{\}$, $\backslash \text{textit}\{\}$ or other formatting that is in the original LaTeX text. 
    
    6. DO NOT hallucinate additional text that is not in the LaTeX source. \\
    
    Additionally, BEFORE you output the latex excerpt in the above format, ABOVE that, I want the category of error that you have identified. \\
    
    Here are all the possible categories of errors:\\
    
    1. Error in algorithm/proof
    
    2. Error in reported results (eg., compared to another paper or original paper)
    
    3. Error in implementation (eg., wrong concept/algorithm used, mismatch between theory and experiments or training and inference, plotting procedure, training procedure)
    
    4. Inconsistencies in definitions
    
    5. Error in assumptions (core conceptual)
    
    6. Incorrect or incomplete analysis
\end{tcolorbox}

\subsection{Internal-Identification}
\begin{tcolorbox}[breakable, colback=black!5!white,colframe=black!20!white,title=\textbf{\textcolor{black}{Internal-Identification Prompt}}]\label{appendix:self_identification}
\small
    Attached is the latex source of a research paper. I have taken a claim in the paper and tried to break it by modifying one or more snippets in the original latex source. You have been provided the modified latex source. Your task is to identify error text chunks that pertain to the MOST serious errors in the paper, and return them in ranked order where the first returned error text are the most serious. These should be substantial technical flaws that could undermine the paper's validity and potentially lead to rejection from a scientific conference or journal.\\

    Important constraints:\\
    - Do not identify minor issues such as grammar, style, formatting, or typographical errors.\\
    - Only use knowledge available up to the time of the paper's publication. Do not search the web or use information published afterward.\\
    - Return exact excerpts from the paper (do not paraphrase).\\
    - Each excerpt must be at most \{word limit\} words long.\\
    - Return at most 10 error chunks, ranked in order of seriousness (most serious first).\\
    
    Output format (strictly follow):\\
    
    :error text:\\
    $<$exact excerpt from the source paper$>$\\
    
    :error text:\\
    $<$exact excerpt from the source paper$>$\\
    
    (...there can be a varying number of error texts for each error, the total number of error text should be at most 10)\\
    
    :explanation:\\
    $<$clear and precise explanation of the error$>$
\end{tcolorbox}

\subsection{Error Identification}
\begin{tcolorbox}
[breakable, colback=black!5!white,colframe=black!20!white,title=\textbf{\textcolor{black}{Error Identification Prompt}}]\label{appendix:error_identification}
\small
    Attached is the PDF of a research paper. Your task is to identify error text chunks that pertain to the MOST serious errors in the paper, and return them in ranked order where the first returned error text are the most serious. These should be substantial technical flaws that could undermine the paper's validity and potentially lead to rejection from a scientific conference or journal.\\
    \\
    Important constraints:\\
    - Do not identify minor issues such as grammar, style, formatting, or typographical errors.\\
    - Only use knowledge available up to the time of the paper's publication. Do not search the web or use information published afterward.\\
    - Return exact excerpts from the paper (do not paraphrase).\\
    - Each excerpt must be at most $\{$word limit$\}$ words long.\\
    - Return at most 10 error chunks, ranked in order of seriousness (most serious first).\\
    \\
    Output format (strictly follow):\\
    \\
    :error-text:\\
    $<$exact excerpt from the source paper$>$\\
    \\
    :error-text:\\
    $<$exact excerpt from the source paper$>$\\
    \\
    (...there can be a varying number of error texts for each error, the total number of error text should be at most 10)\\
    \\
    :explanation:\\
    $<$clear and precise explanation of the error$>$
\end{tcolorbox}

\subsection{Identification Evaluation}
\begin{tcolorbox}
[breakable, colback=black!5!white,colframe=black!20!white,title=\textbf{\textcolor{black}{Identification Evaluation Prompt}}]\label{appendix:error_evaluation}
\small
    I have created a machine learning model to identify errors in a research paper. It identifies the error by pointing to a specific excerpt of the research paper. The output of the model I am going to provide you with is the list of identified errors.\\
    \\
    I also have some ground truth errors. These are the actual errors in the research paper. For this, I am going to provide you with a list of ground truth errors which are again excerpts from the research paper.\\
    \\
    Note that the model need not output errors in the same order as the ground truth errors. Your job is to check each of the identified errors, and see if there is a corresponding ground truth error. For an accurate identified error, at least one ground truth error must be from the same excerpt as the identified error (i.e., there must be at least one match). \\
    \\
    Criteria for a match:\\
    The excerpts of the identified and ground truth errors should be evident that they are the same excerpt from the paper. It is possible that one may be in latex source code and one is plain text, etc. Either the identified error must be a subset of at least one of the ground truth errors OR at least one of the ground truth errors must be a subset of the identified error.\\
    \\
    Provide the output in the following manner:\\
    For each identified error, provide the identified error and then add ``CORRECTLY IDENTIFIED" if it is a correctly identified error (i.e., there is a match with at least one ground truth error), or add ``INCORRECTLY IDENTIFIED" if there is not match with any ground truth error.\\
    \\
    DO NOT EVALUATE the correctness of the error on your own. The correctness is purely determined based on the given ground truth errors. They HAVE to be the same excerpt of the paper.
    \\
    \\IDENTIFIED ERRORS:\\
    $<$identified error texts$>$\\
    \\
    ACTUAL ERRORS:\\
    $<$ground truth error text$>$\\
\end{tcolorbox}

\section{Determining the Evaluation Metric}
\label{appendix:preliminary_eval_exp}

To assess what evaluation metric best aligns with human judgment, we use a small dataset of 20 peer-reviewed papers that are not included in our benchmark. For each paper, we generate modified text that introduces an error using \gpt{} and \gemini{}. We then employ five distinct LLMs --- \texttt{Claude Sonnet 3.7}, \deepseek{},  \gemini{}, \gpt{}, and \grok{} --- to identify the errors in these modified papers. In these examples, we consider the top ten (or less) error excerpts generated by the identification LLM and see whether an error has been identified by using (i) the word-level Levenshtein distance (see Section \ref{section:levenshtein_distance}) (ii) the outputs generated by prompting an instance of the insertion LLM (specific prompt in Appendix \ref{appendix:error_evaluation})(iii) manual inspection, to identify whether any of the top 10 identified error excerpts correspond to the inserted error. We use the manual inspection results as ground truth and compare how the automated methods perform with respect to them. 

Out of the 20 papers, we first manually inspected the inserted errors and found that both insertion models generated valid errors for 19 of the papers. This resulted in 190 verified error instances (19 papers $\times$ 2 insertion models $\times$ 5 identification models). We evaluate automated error identification against manual annotations using two complementary approaches: a Levenshtein-distance-based matching metric and an LLM-based classifier. The results are reported in Table \ref{tab:leven_evaluation_matrix} and Table \ref{tab:llm_evaluation_matrix}.

Across the 190 error instances, we observe that 167 cases (88\%) receive consistent predictions from both the Levenshtein-based method and the LLM-based method, indicating strong agreement between the two signals. While the Levenshtein metric provides a strong and deterministic measure of alignment with the ground-truth annotations, we additionally incorporate the LLM-based predictions to improve recall. Empirically, this combination yields the strongest agreement with manual evaluation, as shown in Table \ref{tab:llm_or_leven_eval_matrix}, where an error is considered identified if either method signals its presence. Accordingly, we adopt this combined criterion as our final evaluation metric, as it best reflects human annotation behavior while maintaining robustness to both surface-level variation and semantic reformulations.

\begin{table}[th!]
\centering
\begin{tabular}{lcccc}
\hline
Label & Precision & Recall & F1 & Support \\
\hline
Not Identified & 0.94 & 0.90 & 0.92 & 148 \\
Identified & 0.69 & 0.79 & 0.73 & 42 \\
\hline
Accuracy & - & - & 0.88 & 190 \\
\hline
\end{tabular}
\caption{Levenshtein vs manual (ground truth) internal error. }
\label{tab:leven_evaluation_matrix}
\end{table}

\begin{table}[th!]
\centering
\begin{tabular}{lcccc}
\hline
Label & Precision & Recall & F1 & Support \\
\hline
Not Identified & 0.89 & 0.90 & 0.90 & 148 \\
Identified & 0.63 & 0.62 & 0.63 & 42 \\
\hline
Accuracy & - & - & 0.84 & 190 \\
\hline
\end{tabular}
\caption{LLM vs manual (ground truth) error identification. }
\label{tab:llm_evaluation_matrix}
\end{table}

\begin{table}[th!]
\centering
\begin{tabular}{lcccc}
\hline
Label & Precision & Recall & F1 & Support \\
\hline
Not Identified & 0.97 & 0.88 & 0.92 & 148 \\
Identified & 0.68 & 0.90 & 0.78 & 42 \\
\hline
Accuracy & - & - & 0.88 & 190 \\
\hline
\end{tabular}
\caption{LLM OR Levenshtein vs manual (ground truth) error identification. }
\label{tab:llm_or_leven_eval_matrix}
\end{table}

\section{Inter-Annotator Agreement Details} 
\label{appendix:inter_annotator_agreement}

Table~\ref{tab:inter_annotator} summarizes agreement statistics for human annotation on LLM error identification results. Across the 285 error identification instances, annotators disagreed on 32 cases (11\%). 
\begin{table}[th!]
    \centering
    \begin{tabular}{ccccc}
    \hline
    \textbf{Insertion Model} & \textbf{Both Identified} & \textbf{Both Not Identified} & \textbf{Disagreement} & \textbf{Total} \\
    \hline
    \gemini{} & 41 & 86 & 18 & $29 \times 5 = 145$ \\
    \gpt{} & 22 & 104 & 14 & $28 \times 5 = 140$ \\
    \hline
    \end{tabular}
    \caption{Inter-annotator agreement for error identification judgments.}
    \label{tab:inter_annotator}
\end{table}

\section{Error Type Analysis}
\label{appendix:error_type}
\subsection{Error Distribution}
\begin{table}[h!]

\centering
\small
\setlength{\tabcolsep}{3pt}
\begin{tabular}{lcc|cc|c}
\hline
 & \multicolumn{5}{c}{\textbf{\gemini{}}} \\
\textbf{Error Type} & Valid Errors & Easy Prompt & Hallucination & Internal ID & Benchmark \\
\hline
Error in Assumptions            & 1906 (99.7\%) & 1659 (87.0\%) & 1476 (89.0\%) & 831 (56.3\%) & 257 (30.9\%) \\
Error in Algorithm/Proof        & 1349 (99.3\%) & 1122 (83.2\%) &  960 (85.6\%) & 378 (39.4\%) & 106 (28.0\%) \\
Incorrect or Incomplete Analysis&  512 (100.0\%)&  383 (74.8\%) &  353 (92.2\%) & 235 (66.6\%) &  54 (23.0\%) \\
Error in Implementation         &  200 (100.0\%)&  179 (89.5\%) &  157 (87.7\%) & 105 (66.9\%) &  26 (24.8\%) \\
Inconsistencies in Definitions  &   23 (95.8\%) &   17 (73.9\%) &   15 (88.2\%) &  10 (66.7\%) &   5 (50.0\%) \\
Error in Reported Results       &   18 (94.7\%) &   13 (72.2\%) &   10 (76.9\%) &   1 (10.0\%) &   0 (0.0\%) \\
\hline
Total & 4008 (99.6\%) & 3373 (84.2\%) & 2971 (88.1\%) & 1650 (55.5\%) & 448 (27.2\%) \\
\hline
\end{tabular}

\caption{Error distribution of \gemini{}-generated errors after each benchmark construction step. The percentages indicate how many errors remain at each step, expressed as a proportion of the errors from the previous stage in the pipeline.}
\label{tab:error_distribution_gemini}
\end{table}

\begin{table}[h!]

\centering
\small
\setlength{\tabcolsep}{3pt}
\begin{tabular}{lcc|cc|c}
\hline
 & \multicolumn{5}{c}{\textbf{\gpt{}}}\\
\textbf{Error Type} & Valid Errors & Easy Prompt & Hallucination & Internal ID & Benchmark \\
\hline
Error in Assumptions              & 3137 (99.4\%)  & 779 (24.8\%) & 614 (78.8\%) & 229 (37.3\%)& 104 (45.4\%) \\
Error in Algorithm/Proof        & 2318 (99.5\%)  & 593 (25.6\%) & 459(77.4\%) & 214 (46.6\%) & 75 (35.0\%)\\
Incorrect or Incomplete Analysis  & 1110 (99.7\%) & 171 (15.4\%)& 139 (81.3\%) & 92 (66.2\%)& 43 (46.7\%)\\
Error in Implementation          &  2186 (99.6\%)& 96 (4.4\%)& 74 (77.1\%)& 45 (60.8\%)& 16 (35.6\%)\\
Inconsistencies in Definitions    &  76 (98.7\%)& 21 (27.6\%) & 16  (76.2\%)& 13 (81.2\%) & 5 (38.5\%)\\
Error in Reported Results         & 218 (98.6\%) & 48 (22.0\%)& 44 (91.7\%)& 34 (77.3\%) & 19  (55.9\%)\\
Unknown                          & \textcolor{lightgray}{60 (98.3\%)}  & \textcolor{lightgray}{13 (21.7\%)} & \textcolor{lightgray}{11 (84.6\%)} & \textcolor{lightgray}{4 (36.4\%)} & \textcolor{lightgray}{3 (75.0\%)} \\
\hline
Total                           & 9105 (99.6\%) & 1722 (18.9\%) & 1358 (78.9\%) & 632 (46.5\%) & 265 (41.9\%) \\
\hline
\end{tabular}
\caption{Error distribution of \gpt{}-generated errors after each benchmark construction step. The percentages indicate how many errors remain at each step, expressed as a proportion of the errors from the previous stage in the pipeline. The ``Unknown" error type are cases when \gpt{} failed to categorize the errors into any of the 6 provided categories.}
\label{tab:error_distribution_openai}
\end{table}

\noindent We analyze the error type distribution after each stage of the benchmark construction. In particular, we obtain the error type by asking the insertion model (\gpt{} for \gpt{}-inserted errors and \gemini{} for \gemini{}-inserted errors) to categorize the generated errors into six distinct categories. This is done by prompting the model with the modified passage, the original passage, the explanation, and the claim. Table \ref{tab:error_distribution_openai} and Table \ref{tab:error_distribution_gemini} present the number of errors in each category at each step of the pipeline.

For errors inserted with \gemini{}, the filtering process is more gradual and distributed across stages. The initial invalid filter removes very few errors in all categories, indicating that most generated errors are considered valid. The easy prompt stage also results in relatively modest reductions, typically removing only around 10–25\% of errors depending on the category. Slightly larger reductions are observed for \textit{incorrect or incomplete analysis} and \textit{inconsistencies in definitions}. The most substantial filtration occurs during the internal identification stage, where reductions become much more pronounced and category-dependent. In particular, \textit{error in algorithm/proof} and \textit{error in reported results} experience the largest drops, with the latter being reduced especially aggressively. 

In contrast, the \gpt{} pipeline exhibits a different behavior, with the majority of filtering occurring at the easy prompt stage. Similar to \gemini{}, the invalid filter removes very few errors. However, during the easy prompt stage, there is a drastic reduction across all categories, with only 4\% to 28\% of errors remaining. This step is particularly aggressive for \textit{error in implementation}, where nearly all errors are eliminated. Compared to this, the internal identification stage is more moderate and uneven across categories: while it continues to reduce errors, the magnitude of reduction is smaller and varies by type, with some categories such as \textit{incorrect or incomplete analysis} and \textit{inconsistencies in definitions} showing relatively limited additional filtering.

These results show that the \gpt{} and \gemini{} pipelines differ not only in how much they filter, but also in when and which categories are most affected. The \gpt{} pipeline  filters early, while \gemini{} applies more substantial filters later, particularly during internal identification. Furthermore, as these reductions vary by category, the composition of errors shifts across stages, and the final benchmark reflects the subset of errors that remain after filtering rather than the initial distribution. This has direct implications for evaluation: overall performance may depend in part on which error types are retained. We therefore analyze model performance by error type to better understand how performance varies across categories that are differentially preserved in the pipeline.

\subsection{Identification Accuracy by Error Type}
Across error types, we observe broadly consistent performance patterns among models. We report results for three error types in Tables \ref{tab:error_in_assumption}–\ref{tab:incorrect_analysis}, as the remaining categories contain too few samples to support stable estimates. Within each reported category, \gpt{} attains the highest estimated identification capability, followed by \deepseek{}, with \claude{}, \grok{}, and \gemini{} occupying lower ranks that vary somewhat across categories. Overall, while larger sample sizes would yield more precise estimates, the observed results indicate that model rankings are relatively stable across the analyzed error types.

\begin{table}[h]
    \centering
    \begin{tabular}{lccc}
    \hline
    \textbf{Identification Model} & \textbf{$\beta_j$} & \textbf{95\% CI} & \textbf{Rank} \\
    \hline
        \gpt{} & 1.68 & [1.19, 2.31] & 1\\
        \deepseek{} & 1.67 & [1.32, 2.22] & 2\\
        \grok{} & 1.30 & [0.92, 1.89] & 4\\
        \claude{} & 1.32 & [1.10, 1.57] & 3\\
        \gemini{} & 1.18 & [0.27, 1.91] & 5\\
    \hline
    \end{tabular}
    \caption{Error identification capabilities for \textbf{Errors in Assumption} for 257 \gemini{} and 104 \gpt{} errors.}
    \label{tab:error_in_assumption}
\end{table}

\begin{table}[h]
    \centering
    \begin{tabular}{lccc}
    \hline
    \textbf{Identification Model} & \textbf{$\beta_j$} & \textbf{95\% CI} & \textbf{Rank} \\
    \hline
        \gpt{} & 2.17 & [1.81, 2.68] & 1\\
        \deepseek{} & 1.42 & [1.03, 1.65] & 3\\
        \grok{} & 1.51 & [1.14, 1.91] & 2\\
        \claude{} & 1.23 & [1.13, 1.44] & 4\\
        \gemini{} & 1.03 & [0.27, 1.65] & 5\\
    \hline
    \end{tabular}
    \caption{Error identification capabilities for \textbf{Errors in Algorithm/Proof} for 106 \gemini{} and 75 \gpt{} errors.}
    \label{tab:error_in_proof}
\end{table}

\begin{table}[h]
    \centering
    \begin{tabular}{lccc}
    \hline
    \textbf{Identification Model} & \textbf{$\beta_j$} & \textbf{95\% CI} & \textbf{Rank} \\
    \hline
        \gpt{} & 6.59 & [3.13, 9.77] & 1\\
        \deepseek{} & 5.58 & [3.74, 7.33] & 2\\
        \grok{} & 4.77 & [2.72, 6.48] & 3\\
        \claude{} & 4.29 & [2.90, 5.61] & 5\\
        \gemini{} & 4.68 & [2.93, 6.97] & 4\\
    \hline
    \end{tabular}
    \caption{Error identification capabilities for \textbf{Incorrect or Incomplete Analysis} for 54 \gemini{} and 43 \gpt{} errors.}
    \label{tab:incorrect_analysis}
\end{table}

\section{Error by Section}
\label{sec:error_distr_sec}

\begin{table}[h!]

\centering
\small
\setlength{\tabcolsep}{2pt}
\begin{tabular}{lcc|cc|c}
\hline
 & \multicolumn{5}{c}{\textbf{\gemini{}}} \\
\textbf{Category} & Valid Errors & Easy Prompt & Hallucination & Internal ID & Benchmark \\
\hline
Methods                           & 1702 (99.3\%) & 1447 (85.0\%) & 1273 (88.0\%) & 628 (49.3\%) & 185 (29.5\%) \\
Exp/Results                       &  664 (99.7\%) &  531 (80.0\%) &  469 (88.3\%) & 311 (66.3\%) &  82 (26.4\%) \\
Exp/Results + Methods             &  572 (99.8\%) &  496 (86.7\%) &  451 (90.9\%) & 226 (50.1\%) &  69 (30.5\%) \\
Intro + Methods                   &  335 (100.0\%) &  285 (85.1\%) &  250 (87.7\%) & 118 (47.2\%) &  32 (27.1\%) \\
Exp/Results + Intro               &  134 (100.0\%) &  112 (83.6\%) &  104 (92.9\%) &  54 (51.9\%) &  11 (20.4\%) \\
Exp/Results + Intro + Methods     &  115 (99.1\%) &   97 (84.3\%) &   86 (88.7\%) &  43 (50.0\%) &  13 (30.2\%) \\
Lit Review                        &   62 (98.4\%) &   48 (77.4\%) &   45 (93.8\%) &  29 (64.4\%) &  11 (37.9\%) \\
Lit Review + Methods              &   61 (100.0\%) &   53 (86.9\%) &   42 (79.2\%) &  21 (50.0\%) &   6 (28.6\%) \\
Intro                             &   52 (100.0\%) &   43 (82.7\%) &   40 (93.0\%) &  16 (40.0\%) &   4 (25.0\%) \\
Exp/Results + Lit Review          &   31 (100.0\%) &   28 (90.3\%) &   24 (85.7\%) &  13 (54.2\%) &   4 (30.8\%) \\
Other Combinations                &  280 (100.0\%) &  233 (83.2\%) &  187 (80.3\%) & 101 (54.0\%) &  31 (30.7\%) \\
\hline
Total                             & 4008 (99.6\%) & 3373 (84.2\%) & 2971 (88.1\%) & 1560 (52.5\%) & 448 (28.7\%) \\
\hline
\end{tabular}
\caption{Section combination distribution of \gemini{}-generated errors after each benchmark construction step. The percentages indicate how many errors remain at each step, expressed as a proportion of the errors from the previous stage in the pipeline.}
\label{tab:section_combos_gemini}
\end{table}

\begin{table}[h!]

\centering
\small
\setlength{\tabcolsep}{2pt}
\begin{tabular}{lcc|cc|c}
\hline
 & \multicolumn{5}{c}{\textbf{\gpt{}}} \\
\textbf{Category} & Valid Errors & Easy Filter & Hallucination & Internal ID & Benchmark \\
\hline
Methods                       & 2643 (99.4\%) & 690 (26.1\%) & 541 (78.4\%) & 238 (44.0\%) &  92 (38.7\%) \\
Exp/Results                   & 2426 (99.5\%) & 328 (13.5\%) & 257 (78.4\%) & 157 (61.1\%) &  74 (47.1\%) \\
Exp/Results + Methods         & 2270 (99.5\%) & 367 (16.2\%) & 296 (80.7\%) & 128 (43.2\%) &  53 (41.4\%) \\
Intro + Methods               &  382 (99.5\%) &  73 (19.1\%) &  58 (79.5\%) &  21 (36.2\%) &   7 (33.3\%) \\
Exp/Results + Intro           &  199 (99.0\%) &  35 (17.6\%) &  26 (74.3\%) &  17 (65.4\%) &  10 (58.8\%) \\
Exp/Results + Intro + Methods &  220 (99.5\%) &  37 (16.8\%) &  25 (67.6\%) &   7 (28.0\%) &   2 (28.6\%) \\
Lit Review                    &   70 (100.0\%) &  24 (34.3\%) &  23 (95.8\%) &  11 (47.8\%) &   5 (45.5\%) \\
Lit Review + Methods          &  148 (100.0\%) &  32 (21.6\%) &  24 (75.0\%) &   9 (37.5\%) &   6 (66.7\%) \\
Intro                         &   82 (100.0\%) &  19 (23.2\%) &  17 (89.5\%) &   8 (47.1\%) &   2 (25.0\%) \\
Exp/Results + Lit Review      &  110 (100.0\%) &  21 (19.1\%) &  17 (81.0\%) &   7 (41.2\%) &   2 (28.6\%) \\
Other Combinations            &  555 (99.1\%) &  96 (17.3\%) &  74 (77.1\%) &  29 (39.2\%) &  12 (41.4\%) \\
\hline
Total                         & 9105 (99.5\%) & 1722 (18.9\%) & 1358 (78.9\%) & 632 (46.5\%) & 265 (41.9\%) \\
\hline
\end{tabular}

\caption{Section combination distribution of \gpt{}-generated errors after each benchmark construction step. The percentages indicate how many errors remain at each step, expressed as a proportion of the errors from the previous stage in the pipeline.}
\label{tab:section_combos_openai}
\end{table}

To further understand the categorization of errors that progressed through the various filtering stages, we analyze the section of the paper in which the error was present. Since a single error in a paper can result from the modification of more than one excerpt, it is possible that these excerpts lie in different sections of the paper. To categorize an error by section, we first identify the section that each excerpt that constitutes the error belongs to. We then broadly classify each section into one of \textit{Introduction (Intro)}, \textit{Literature Review (Lit Review)}, \textit{Methodology (Methods)}, \textit{Experiments, Results and Discussion (Exp/Results)}, \textit{Conclusion}, and \textit{Other}. If an error results from the modification of excerpts in more than one section, we report the combination of sections through the use of the ``+'' icon in Tables~\ref{tab:section_combos_gemini} and \ref{tab:section_combos_openai}, which provide the number of errors in the top 10 section combinations at each filtering stage, generated by \gemini{} and \gpt{} respectively. 

For \gemini{} generated errors, the majority of errors span the \textit{Methodology} section (1,702 valid errors), followed by \textit{Experiments, Results and Discussion} and \textit{Methodology + Experiments, Results and Discussion} combinations. Across the full filtering pipeline (valid errors through to internal identification), errors located purely in the \textit{Introduction} section are filtered most aggressively, with only 30.8\% surviving, followed by errors in \textit{Literature Review + Methodology} (34.4\%) and \textit{Introduction + Methodology} (35.2\%). In contrast, errors in \textit{Experiments, Results and Discussion} alone and \textit{Literature Review} alone show the highest survival rates (46.8\% each), suggesting these tend to be subtler and harder to detect.

For \gpt{} generated errors, the filtering is far more aggressive overall, with survival rates from valid errors to internal identification in the single digits for most section combinations. Errors spanning \textit{Introduction + Methodology + Experiments, Results and Discussion} are filtered most heavily, with only 3.2\% surviving to internal identification, followed by \textit{Introduction + Methodology} (5.5\%) and \textit{Experiments, Results and Discussion + Methodology} (5.6\%). Errors confined to \textit{Literature Review} alone survive at the highest rate (15.7\%), mirroring the Gemini trend.

On manually observing some errors before and after the filtering process, we find that errors are largely filtered out due to the following four reasons. The first is that they are mathematically trivial or algebraically degenerate errors. The second is that there exists some direct internal contradiction. The third consists of an altered methodology which is obviously infeasible, and the fourth is due to the contradiction of well-known domain principles. 

\section{Extension of \benchmark{} to Non-Machine Learning Domain}
\label{appendix:extension}
The main results in this paper have been obtained by applying our pipeline on papers from ICML 2025, a premier machine learning conference. To further assess whether our error insertion pipeline and identification task generalize beyond this domain, we additionally evaluate it on papers outside of machine learning. To this end, we collected 100 papers from five different conferences, which are within the field of computer science but not machine learning. The five conferences are the International Conference on Architectural Support for Programming Languages and Operating Systems (ASPLOS), The ACM Conference on Computer and Communications Security (CCS), the International Conference on Software Engineering (ICSE), the ACM SIGPLAN Conference on Programming Language Design and Implementation (PLDI), and the IEEE Symposium on Security and Privacy.

Table~\ref{tab:filtering_papers_nonML} summarizes the number of unique papers and injected errors after each stage of the pipeline. Similar to the main benchmark, we observe substantial filtering of \gpt{}-inserted errors, with only 7\% retained, whereas \gemini{}-inserted errors are preserved at a higher rate (32\%). From the subset of papers whose PDFs compile successfully, we sample one error per paper, resulting in a final extension set of $67 + 48 = 115$ errors.  To evaluate insertion quality, we manually annotate a subset of 24 errors, consisting of 12 \gpt{}-inserted and 12 \gemini{}-inserted errors on the same set of 12 papers. Each error is independently reviewed by two expert human annotators. The annotators disagree on only one instance, corresponding to an agreement rate of 96\%.

Table~\ref{tab:model_ranking_nonML} reports model identification accuracy on this dataset. Overall, performance trends are consistent with those observed on ICML papers. At $k=10$, the relative ranking of models remains unchanged, with \gpt{} achieving the highest performance. At $k=3$, however, \deepseek{} and \grok{} outperform \gpt{}. The non-ML papers are drawn from January 2024 onward. As such, there is a possibility of data leakage if identification models have previously seen these papers. Despite this, performance across all models remains low, still with no model surpassing 50\% identification accuracy even at $k=10$. Finally, we note that this extension is limited by its relatively small sample size and should therefore be interpreted as a preliminary validation of the pipeline’s generalizability, rather than a definitive conclusion. Nevertheless, the results suggest that the observed trends are not specific to machine learning papers. 

\begin{table}[t]
    \centering
    \begin{tabular}{p{4.6cm}cccc}
    \hline
    
    \multirow{2}{*}{\textbf{Insertion Model}} & \multicolumn{2}{c}{\textbf{\gemini{}}}& \multicolumn{2}{c}{\textbf{\gpt{}}} \\

    & \# Papers & \# Errors & \# Papers & \# Errors\\
    \hline
     
    Claim Extraction & 100 & 742 & 100 & 1735 \\
    Error Generation & 100 (100\%) & 742 & 99 (99\%) & 1735 \\
    \hline
    Filtering Invalid Errors & 100 (100\%) & 741 (100\%) & 99 (100\%) & 1726 (99\%)\\
    Filtering Easy Errors Stage 1, \textit{Easy Error Prompt Filter} & 100 (100\%) & 604 (81\%) & 84 (85\%) & 299 (17\%)\\
    Filtering Easy Errors Stage 2, \textit{Internal Identification Filter} & 98 (98\%) & 385 (52\%) & 69 (70\%) & 163 (9\%)\\
    \hline
    PDF Compilation & 67 (67\%) & 241 (32\%) & 48 (48\%) & 113 (7\%)\\
    \hline
    \end{tabular}
    \caption{Number of unique papers and errors after each step in the error insertion pipeline for non-Machine Learning papers. The percentages are based on the initial number of papers and errors. }
    \label{tab:filtering_papers_nonML}
    
\end{table}

\begin{table}[t]
    \centering
    \begin{tabular}{lcc}
    \hline 
    \textbf{Identification Model} & \textbf{Accuracy @k=3} & \textbf{Accuracy @k=10}\\
    \hline
    \gpt{} & 17.9\% & 46.3\%\\
    \deepseek{} & 23.5\% & 27.8\%\\
    \grok{} & 19.1\% & 27.0\%\\
    \claude{} & 14.8\% & 21.7\%\\
    \gemini{} & 12.5\% & 14.6\%\\
    \hline
    \end{tabular}
    \caption{Ranking of LLM performance on \benchmark{} created using non-Machine Learning papers. The ``Accuracy'' columns correspond to the fraction of instances where the top-ranked $k$ error candidate texts output by the LLM actually contain the error.}
    \label{tab:model_ranking_nonML}
\end{table}
\end{document}